\pdfoutput=1
\documentclass[11pt]{article}

\usepackage{acl}

\usepackage{times}
\usepackage{latexsym}

\usepackage{microtype}
\usepackage{microtype}
\usepackage{amsfonts}
\usepackage{amssymb}
\usepackage{bbold}
\usepackage{graphicx}
\usepackage{multirow}
\usepackage{xcolor}
\usepackage[hang,flushmargin]{footmisc} 
\usepackage{dsfont}
\usepackage{upgreek}
\usepackage{amsmath}
\usepackage{nccmath}
\usepackage{mathtools}
\usepackage{devanagari}
\newcommand\blfootnote[1]{%
  \begingroup
  \renewcommand\thefootnote{}\footnote{#1}%
  \addtocounter{footnote}{-1}%
  \endgroup
}

\usepackage[T1]{fontenc}
\usepackage[utf8]{inputenc}

\usepackage{microtype}

\title{Denoising Large-Scale Image Captioning from \\ Alt-text Data using Content Selection Models}

\author {
        \textbf{Khyathi Raghavi Chandu}\textsuperscript{\rm $\blacklozenge$ \textbf{1}}, 
        \textbf{Piyush Sharma}\textsuperscript{\rm $\clubsuit$ \textbf{2}},\\
        \textbf{Soravit Changpinyo}\textsuperscript{\rm $\spadesuit$}, \textbf{Ashish Thapliyal}\textsuperscript{\rm $\spadesuit$}, \textbf{Radu Soricut}\textsuperscript{\rm $\spadesuit$} \\ \\
        \textsuperscript{\rm $\blacklozenge$} Meta AI, \textsuperscript{\rm $\clubsuit$} Uber, 
        \textsuperscript{\rm $\spadesuit$} Google Research \\ 
        \texttt{\small {khyathi.research@gmail.com, \{schangpi, asht, rsoricut\}@google.com}}
}

\begin{document}
\maketitle
\blfootnote{\textbf{1} Work done during internship at Google Research}
\blfootnote{\textbf{2} Work done while employed at Google Research}

\begin{abstract}
Training large-scale image captioning (IC) models demands access to a rich and diverse set of training examples that are expensive to curate both in terms of time and man-power. 
Instead, using alt-text based captions gathered from the web is a far cheaper alternative for scaling with the downside being that the data is noisy.
Recent modeling approaches to IC often fall short in terms of performance in leveraging these noisy datasets as compared to datasets with clean  annotations.
%this case, because they
%assume a clean annotated dataset (as opposed to the noisier alt-text--based annotations), and
% employ an end-to-end generation approach, which often lacks both controllability and interpretability.
We address this problem with a simple yet effective technique of breaking down the task into two smaller, 
more controllable tasks -- skeleton prediction and skeleton-based caption generation. Specifically,
we show that \textit{sub-selecting content words as skeletons} helps in generating improved and denoised captions 
when leveraging rich yet noisy alt-text--based \textit{uncurated} datasets. 
We also show that the predicted English skeletons can further cross-lingually be leveraged to generate non-English captions, by presenting
experimental results in French, Italian, German, Spanish and Hindi. 
We also show that skeleton-based prediction allows for better control of caption properties, such as length, content, and gender expression, providing a handle to perform human-in-the-loop interpretable semi-automatic corrections.
\end{abstract}

\section{Introduction}

In the last demi-decade, NLP fields have ventured into reaping the benefits of utilizing large scale raw \textit{(uncurated)} data from web-crawls. 
This trend aligned with new uncurated image-captioning datasets like Conceptual Captions \cite{DBLP:conf/acl/SoricutDSG18}.
While these uncurated datasets are superior in terms of size and diversity, they are inferior to well curated datasets \cite{DBLP:conf/eccv/LinMBHPRDZ14, wang2019balanced} in terms of noise in the captions. The content in the alt-text for the image is often distorted by the intent or context in which the image is presented. For example, the ground truth alt-text caption for a house is \textit{`house for sale'} instead of \textit{`front view of a house'}. This hinders the use of these large noisy datasets to the fullest extent.

\begin{figure}[t!]
\centering
\includegraphics[trim=0cm 0cm 0cm 0cm, width=\linewidth]{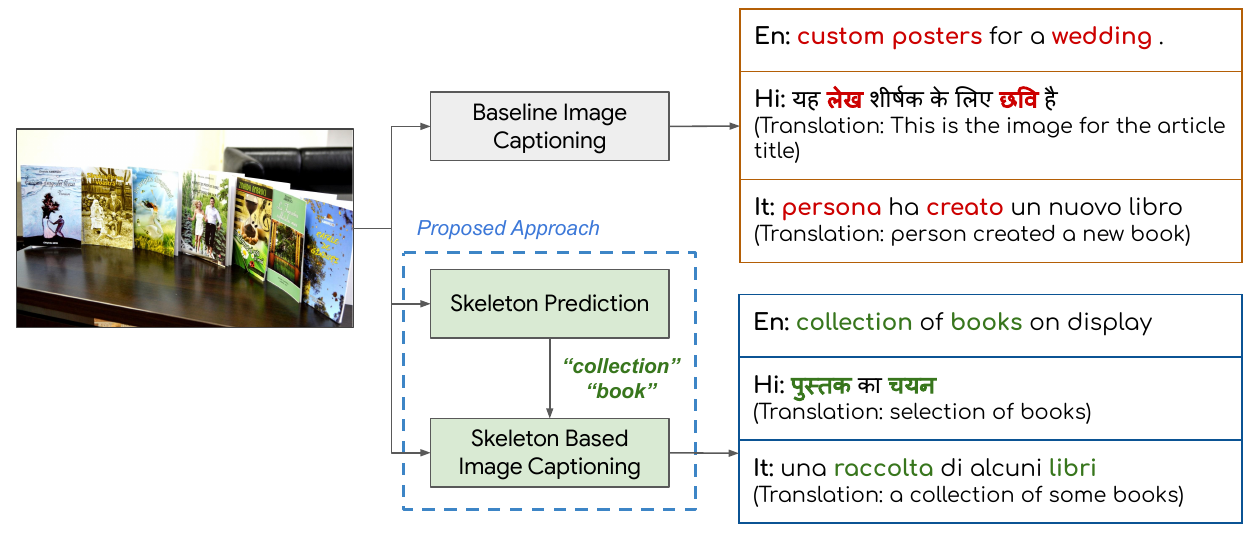}
%\vspace{-6pt}
\caption{ {\small
Overview of our approach: (1) skeleton prediction \&  (2) skeleton based IC; compared to conventional IC. 
Output captions shown in English (En), Hindi (Hi) and   Italian (It).
}}
\vspace{-15pt}
\label{fig:intro}
\end{figure}

We present a simple two-staged approach by separating the content selection from caption generation as illustrated in Figure \ref{fig:intro}. In contrast to most IC approaches \cite{DBLP:journals/corr/abs-1810-04020, sharma2020image}, which hallucinate incorrect content from noisy training data (i.e `custom posters' and `wedding'), our approach first focuses on \textit{denoising} the content words (i.e `collection' and `book') that are further used to generate a relevant caption. We refer to this sequence of concept words that are key pieces of information consistent with the image as a \textit{skeleton}. 
Sub-selecting skeleton words that curb noisiness are automatically extracted from the alt-text captions. 
We focus on language-based skeletons that are derived from captions \cite{kuznetsova2014treetalk,DBLP:conf/cvpr/FangGISDDGHMPZZ15,DBLP:conf/nips/DaiFL18}, rather than expensive visual-based skeletons derived from image, e.g., scene graphs, \cite{wang2019role,DBLP:conf/cvpr/YangTZC19}, which are hard to scale.
More concretely, we introduce an intermediate task of distantly supervised skeleton prediction in the end to end IC pipeline: 
The end-to-end task of IC ($f_{\theta}: {\mathbb{I} \rightarrow \mathbb{C}}$) is broken down
into a two-staged  
pipeline: skeleton prediction ($f_{\theta}:{\mathbb{I}\rightarrow\mathbb{S}}$) and skeleton based captioning ($f_{\phi}:{\mathbb{I, S}\rightarrow\mathbb{C}}$),
where $\mathbb{I}$ is the image, $\mathbb{S}$ is the skeleton, and $\mathbb{C}$ is the caption  \cite{DBLP:conf/cvpr/KulkarniPDLCBB11, li2011composing, DBLP:conf/emnlp/ElliottK13, DBLP:conf/cvpr/FangGISDDGHMPZZ15}. 
We present a comparison between encoding, decoding and autoencoding these skeletons.
As such, our skeleton prediction solution addresses the \textit{semantic gap} problem \cite{DBLP:conf/ijcai/LiC18,DBLP:conf/eccv/YaoPLM18}.

We illustrate the effectiveness of this approach on  uncurated noisy datasets in the following ways.
(1) We demonstrate that sub-selecting content words with an intermediate \textit{skeleton prediction task denoises content} thereby leading to better human evaluation results on captioning. We also conduct an extensive analysis on multimodal discourse relations and find that the reason for this improvement is the generation of more visible captions \cite{alikhani2020cross}.
(2) Scaling large uncurated datasets to other languages is still a bottleneck. We show the \textit{transferability of learning English skeletons} to improve caption generation in other languages -- English, French, Italian, German, Spanish and Hindi. 
(3) The predicted skeletons qualitatively demonstrate other potential benefits, such as \textit{controllability} of content, length, and  %\cite{DBLP:conf/cvpr/CorniaBC19, DBLP:journals/corr/abs-2007-09580},
gender via a natural language--based \textit{interpretable} interface, which enables one to additionally interact with the generation process.

\section{Related Work}
\label{sec:rw}

% Content selection via caption words
% \citet{kuznetsova2014treetalk,DBLP:conf/cvpr/FangGISDDGHMPZZ15,DBLP:conf/nips/DaiFL18}
% Content selection via scene graphs
% \cite{wang2019role,DBLP:conf/cvpr/YangTZC19}

% Past decade has seen massive developments in novel sequence generation tasks with neural models utilizing RNNs and transformers. 
% However, majority of this work operates with a symmetry in training and test distributions. Our work bridges the gap between these distributions using compositional tasks within IC.

\paragraph{Content selection from vision: }
There is a rich body of work in improving content selection for IC \cite{DBLP:conf/cvpr/Feng00L19a}, mainly focused on  scene graph based skeletons \cite{DBLP:conf/iccv/GuJCZYW19, kimvizwiz, DBLP:journals/corr/abs-2003-00387, DBLP:conf/cvpr/YangTZC19}. %While  also rely on expensive scene graphs, they leverage an intermediate dictionary to derive more descriptive captions. 
However, these annotations with objects and relations are expensive, thereby constraining the scaling up to  multiple languages and diverse concepts. Our work delegates this responsibility of identifying content to the language modality by using inexpensive off the shelf tools for weak supervision.

\noindent \textbf{Content selection from language: } An orthogonal body of work relies on skeletons derived from language using hierarchical phrase modeling \cite{DBLP:conf/accv/TanC16, DBLP:conf/nips/DaiFL18}, semantic attention \cite{you2016image}, attribute LSTM \cite{yao2017boosting}, 
%predicting a skeleton first and then filling their attributes 
skeleton based attribute filling 
\cite{wang2017skeleton}, adaptively merging topic and visual information
%step by step 
\cite{liu2018simnet}, multimodal flow \cite{DBLP:conf/iccv/LiZLY19} and concept guided attention \cite{li2019boosted}. Note that all these prior works utilize human curated gold datasets such as COCO \cite{DBLP:conf/eccv/LinMBHPRDZ14} and Flickr30k \cite{plummer2015flickr30k} with clean coupling between captions and images. However, scaling them to large and diverse concepts is expensive. We utilize \textit{uncurated} silver standard datasets with the advantages of richness and diversity at the cost of noisy text. Hence we show the effectiveness of a dual staged approach that denoises the captions by skeleton prediction.

\paragraph{Cross-lingual and controllable captions: }
%One aspect of guiding captions is generating captions in other languages. 
Past work on cross-lingual captioning focused on translation \cite{DBLP:conf/wmt/BarraultBSLEF18}, fluency guidance \cite{lan2017fluency}, using large datasets \cite{DBLP:conf/acl/YoshikawaST17} and more recently by pivoting on source language captions \cite{DBLP:conf/acl/ThapliyalS20, DBLP:conf/eccv/GuJCW18}. We go a step further and pivot on the predicted English skeleton to improve multilingual captions due to the 
dearth of similar off the shelf tools in other languages.
% We utilize this to better transfer the content from one language to another due to the 
% dearth of similar off the shelf tools in other languages.
%demonstrate improvements in multilingual captions by pivoting on the predicted skeleton in English.
%\citet{DBLP:conf/acl/ThapliyalS20} propose a pivot based model for cross lingual IC by pivoting on caption from source language in the decoder. \citet{DBLP:conf/eccv/GuJCW18} perform language pivoting in captioning by reconstructing both the modalities and regularizing the language embeddings across pivoting.
% Specifically, they pivot on language with respect to which there is abundant parallel IC data as well as parallel machine translation data with respect to the target language. 
%Similar to our SkeDecoding model in spirit, 
We qualitatively explore controlling length via skeletons which was explored before via adding length to decoder \cite{DBLP:journals/corr/abs-2005-14386, 
DBLP:conf/cvpr/CorniaBC19}.
%DBLP:journals/corr/abs-2007-09580} 
% and language \cite{tsutsui2017using} can be controlled in the decoder %similar to one of our models.
% which we qualitatively explore. 
Other controllable aspects include stylistic captions \cite{guo2019mscap, mathews2018semstyle}
language  \cite{tsutsui2017using} 
which %we believe 
are potential extensions to our work. 
%In addition to language, other controllable properties include length \citet{DBLP:journals/corr/abs-2005-14386} (similar to SkeDec), 

%In addition, length can also be controlled by injecting the remaining length at each time step of the decoder \citet{DBLP:journals/corr/abs-2005-14386}. More specifically, the control token in this case is the length which is predicted before decoding. Similarly, in our work, we predict the skeleton before decoding the caption. %\cite{DBLP:journals/corr/abs-2004-08070} control the entities to derive informative captions from news articles by modeling objects and faces separately. 
% \citet{DBLP:journals/corr/abs-2003-03107} propose an iterative adaptive refinement in order to regenerate an existing caption with copy mechanism. \cite{ren13hybrid} propose hybrid channels for information control with two levels of decoding: one with self study image features from a transformer, and the second with teacher guided information from ground truth captions.% combined using a gating mechanism. 

\paragraph{Interpretable Natural language skeletons: }
Despite remarkable advancements of large scale end-to-end models, recent work identifies spurious correlations in datasets
%being optimized for maximum likelihood 
that potentially lead to high performance \cite{DBLP:conf/emnlp/GevaGB19, DBLP:conf/lrec/Tsuchiya18}. To mitigate this, researchers began to dissect intermediate components of models with the goal of interpretability to humans \cite{DBLP:conf/emnlp/WiegreffeP19, DBLP:conf/naacl/ThorneVCM19, DBLP:journals/cacm/Lipton18} as opposed to %the body of work based on 
implicit explanation \cite{xu2015show}. Our work is an instance of explaining captions through skeleton predictions %that are weakly supervised 
similar to recent work on rationalizing answer predictions for question answering \cite{DBLP:journals/corr/abs-2004-05569}. %Since generating an entire rationale to assist caption generation might be somewhat redundant, 
We view the intermediate skeleton layer as an interpretable model prediction that helps us study key subtle dataset attributes, such as gender bias.% and provide human-in-the-loop interventions to improve the final caption.

%Our work can also be viewed as an instance of explaining captions through skeleton predictions that are weakly supervised similar to recent works in rationalizing answer predictions through  ra 

\section{Our Approach}
\label{sec:method}

\begin{figure*}[t!]
\centering
\includegraphics[trim=1.5cm 3.5cm 0.2cm 4.6cm, clip=true, width=0.695\linewidth]{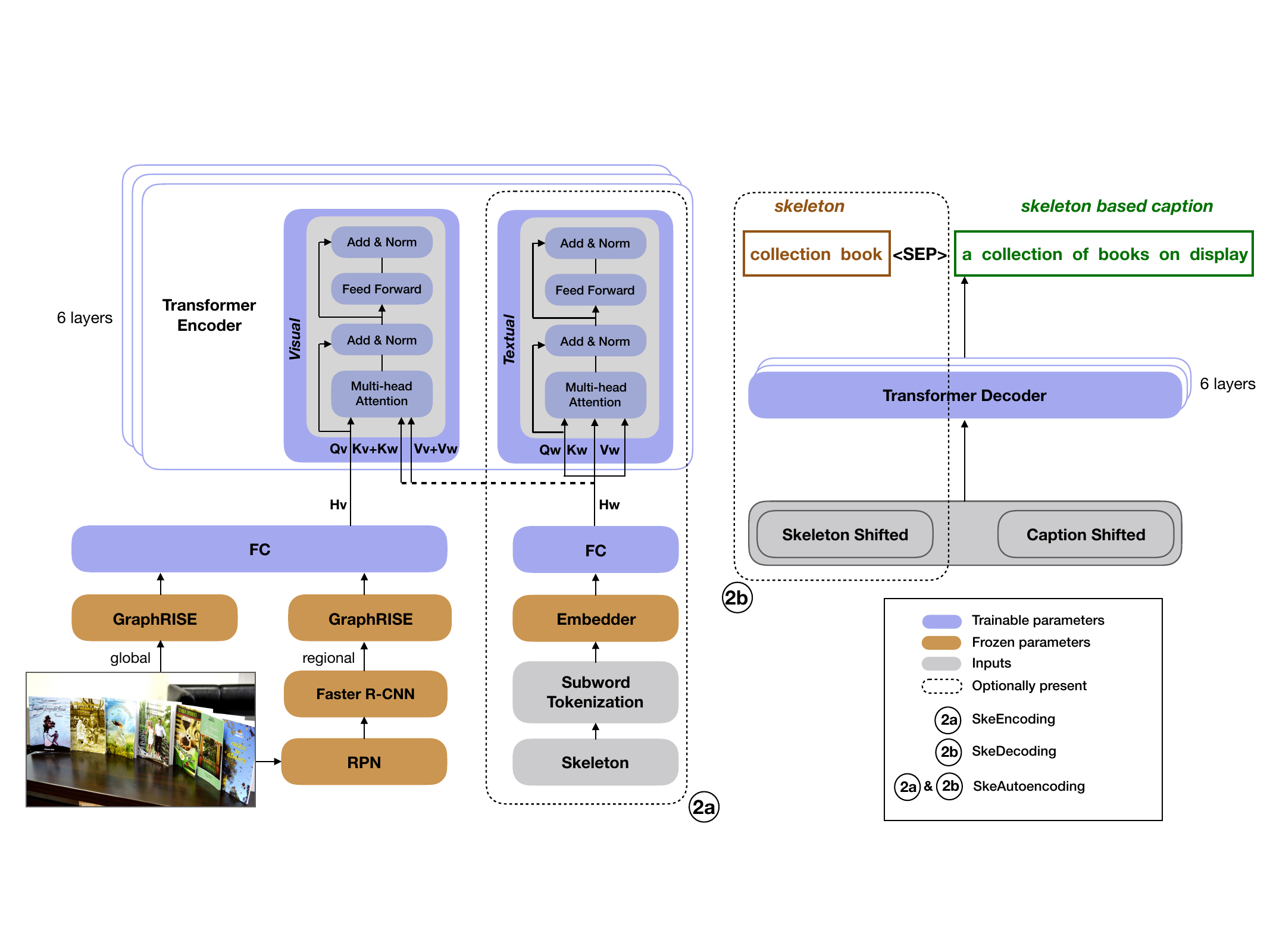}
\vspace{-6pt}
\caption{{\small Model architecture of our skeleton based captioning along with \textit{text as side attention} mechanism between visual (v) and textual (w) modalities. The skeleton is present optionally in the encoder, decoder or both based on our three approaches.} }
%\TODO{Any ideas for writing `optionally present' in a better way?}}
\label{fig:architecture}
%\vspace{-15pt}
\end{figure*}

%IC is the task of generating a descriptive sentence from an image where a neural model expects a bunch of image features that in turn generates a meaningful and relevant target sentence. 
IC requires paired examples of images and captions  ($\mathbb{I}$, $\mathbb{C}$), where $c \in \mathbb{C}$ correspond to tokens in a caption ($c_1, c_2, ..., c_m$), which are often expensive to gather.
Under this paradigm, end-to-end model training attempts to perform a match between the semantic concepts present in $\mathbb{I}$ and $\mathbb{C}$, 
starting from image, region, and object level features and mapping them to various $c_i$s.
In contrast, our approach uses intermediate skeletons as an effective way to leverage noisy, uncurated alt-text based captions to train a model to generate more visually informative captions.
%In contrast, our model bridges the semantic gap by introducing a compositional skeleton in between, which breaks down the task of $f_{\theta}: {\mathbb{I} \rightarrow \mathbb{C}}$ into two tasks $f_{\theta}: {\mathbb{I} \rightarrow \mathbb{S}}$ and $f_{\phi}: {\mathbb{I, S} \rightarrow \mathbb{C}}$. 
%We observe that this decoupling improves performance by denoising semantic content prediction when using large, rich alt-text--based datasets for training, such as Conceptual Captions~\cite{DBLP:conf/acl/SoricutDSG18} (CC).
An overview of both the stages is presented in Fig.~\ref{fig:intro}.

\subsection{Distantly Supervised Skeletons }
\label{sec:skeleton-types}

Since gold standard skeleton words are usually not available, % for IC datasets, 
we use distant supervision to get these labels. We retrieve syntax annotations (POS tags and word lemmas), using the Google Cloud Natural Language API \footnote{\href{https://cloud.google.com/natural-language}{https://cloud.google.com/natural-language}} of caption texts. 
We use these annotations to experiment with skeleton variants. 
The ground-truth skeletons are selected by analyzing the syntax of the automatically curated web-scaled captions through combinations of nouns, verbs, adjectives and adverbs in their condensed forms. In addition, we also ignore tokens with a frequency of less than 50 in our training data to reduce noise while selecting the skeleton words. This subselection of content based on POS tags and downscaling of vocabulary helps in retaining important words as skeletons resulting in a label size of 5k. Since automatic n-gram based metrics cannot be evaluated against noisy ground-truths, manual evaluation is conducted to understand the denoising of sub-selection.

%skeletons with the following four variants.
% We use an oversimplified form of the syntactic relations in the captions to represent 3 skeleton forms.
% The following are the three variants followed by a baseline skeleton form.
% Utilizing these syntax variants, we examine the following skeletal forms. 

%  In the spirit of simplicity, the same model is used to generate a bunch of relevant concepts for the image.

%\begin{enumerate}
\noindent \textit{1. Nouns \& Verbs:} This includes a sequence of lemmas of all the nouns and verbs in a caption.% sentence.

\noindent \textit{2. Salient Nouns \& Verbs:} 
        Saliency of nouns and verbs is determined using tf-idf scores, treating each caption as a document. 
        For each caption, the top 2 highest scoring noun and verb tokens (lemma) are selected.
        This examines if saliency contributes towards effectiveness of the skeleton.

\noindent \textit{3. Nouns:} 
        This includes lemmas of all the nouns. %and we are examining the role of the actions while generating a caption. 
        This helps us untangle the roles of nouns vs verbs in the effectiveness of the skeleton. 

\noindent \textit{4. Iteratively refined captions:} 
        Under this condition, the output of the baseline Img2Cap model serves as the `skeleton' for the next skeleton-based captioning stage.
        The rationale behind this skeleton is to compare the utility of sub-selecting skeleton words based on POS in denoising caption content, compared to a full caption prediction.
        % More concretely, this skeleton investigates if the model is correcting the previously made errors or if the model is doubling down on the same errors.
%\end{enumerate}

%We ignore skeleton tokens with a frequency of less than 50 in our training data to reduce noise. 
%A frequency based pruning with 50 is performed on skeleton words to reduce noise. This POS-tag based subselection of content and downscaling of vocabulary helps in retaining important words as skeletons resulting in a label size of 5k.

% \noindent \textbf{Baseline $f_{\theta}: {\mathbb{I} \rightarrow \mathbb{C}}$ (Img2Cap): } 

\subsection{Model }
\paragraph{Baseline (Img2Cap):} 
We adopt an encoder-decoder ($f_{\theta}: {\mathbb{I} \rightarrow \mathbb{C}}$) IC model based on Transformers~\cite{DBLP:conf/nips/VaswaniSPUJGKP17} following recent state-of-the-art approaches \cite{DBLP:conf/acl/SoricutDSG18,yu2019multimodal,changpinyo2019decoupled,huang19attention,cornia20m2}.
%As a starting point, 
Our model uses the IC framework introduced in~\cite{changpinyo2019decoupled}.
Inspired by the bottom-up and top-down %attention BUTD 
approach~\cite{anderson2018bottomup}, the input image $\mathbb{I}$ is represented as
a bag of features, containing one global and 16 regional, fine-grained feature vectors.
The regional features correspond to the top 16 box proposals from a Faster-RCNN~\cite{ren2015faster} object detector trained on Visual Genome~\cite{krishnavisualgenome}, with a %backbone
ResNet101~\cite{resnet} that is trained on JFT~\cite{hinton2015distilling} and fine-tuned on ImageNet~\cite{imagenet15}. 
We featurize both global and regional boxes using Graph-RISE~\cite{DBLP:journals/corr/abs-1902-10814, DBLP:conf/wsdm/JuanLLPTCGDTR20}.
We make the following %modifications
changes to the state of the art model %from
\cite{changpinyo2019decoupled}, leading to a 9-point improvement on the dev CIDEr %score
on CC (1.00 vs. 0.91) (\textbf{improved baseline}): 1) encode the corners and the area of the bounding boxes to fuse positional information with visual features, %as in
~\cite{lu19vilbert}, and 2) encode each feature vector with a Linear-ReLU-LayerNorm-Linear instead of Linear embedding layer, where LayerNorm is layer normalization~\cite{ba2016layer}.

% \TODO{Khyathi,Beer: Please check these numbers, specifically where 0.91 came from.}
% \khyathi{Is this the G + B-Ultra in the decoupled2019 paper?}
% \TODO{Khyathi,Beer: JFT}
% \beer{Yep. As we do not have ICA labels here.}

% Given , the visual features are extracted using Graph-Regularized Image Semantic Embedding (GraphRISE).

% \noindent \textbf{Dual Staged Composition of Simplified Tasks: }
\paragraph{Dual Staged Modeling:}
In this approach, we introduce an intermediate natural-language interpretable skeleton $\mathbb{S}$ between $\mathbb{I}$ and $\mathbb{C}$.
This $\mathbb{S}$ is composed of a sequence of lemmas, using a subset of content words ($s_1, s_2, ... s_n$) from $c$, where $n<m$. This reduces the output complexity of $f_{\theta}: {\mathbb{I} \rightarrow \mathbb{C}}$ by simplifying and denoising the noisy $\mathbb{C}$ to $\mathbb{S}$. %with the goal of denoising
%to denoise the silver captions.
%. We posit that this simplified representation helps denoise the captions. 
Hence, the task of IC is decomposed into the first stage of predicting skeleton concepts and the second stage of %generating the caption
caption generation %by utilizing
using the intermediate skeleton. 

%\TODO{We claim above that $\mathbb{S}$ is simpler that $\mathbb{C}$. We should explain that. I added something on the lines of ``morphological root words''. We should explain, perhaps with an example.}

\iffalse
{\small
\begin{itemize}
    \item $\mathbb{T}_{is}$: Skeleton Prediction
    \begin{itemize}
        \item Input: Images ($\mathbb{I}_{j}$) 
        \item Output: Skeleton ($\mathbb{S}_{j}$)
    \end{itemize}
    \item $\mathbb{T}_{isc}$: Caption Generation
    \begin{itemize}
        \item Input: Images ($\mathbb{I}_{j}$), Skeleton ($\mathbb{S}_{j}$)
        \item Output: Caption, $\mathbb{C}_{j}$
    \end{itemize}
\end{itemize}
}
\fi

% \noindent \textbf{Stage 1: Skeleton Prediction $f_{\theta}: {\mathbb{I} \rightarrow \mathbb{S}}$ (Img2Ske): }
\paragraph{Stage 1: Skeleton Prediction (Img2Ske): }
The first stage ($f_{\theta}: {\mathbb{I} \rightarrow \mathbb{S}}$) is to predict one of the 4 variants of the skeleton words (from \S \ref{sec:skeleton-types}) from the images. We experiment with both classification and generation paradigm that respectively do not possess and possess linear conditioning of the predicted skeleton word on the following words. We observe that the generation based skeleton prediction results in skeleton words that  co-occur in a sentence. In contrast, the classification approach predicts skeleton words relevant to an image like \textit{{person, man, singer}} that do not necessarily co-occur in a caption. This is detailed in \S \ref{sec:classification} of Appendix.
To improve co-occurrence of the predicted skeleton words, we generate %skeleton
%based approach 
%addresses the co-occurrence problem by predicting 
the skeleton words $\hat{\mathbb{S}}$ autoregressively where each word is conditioned on the previously predicted skeleton word.
This conditional dependence models word co-occurrence more tightly as  $p(\hat{\mathbb{s}}_{j} |I, \hat{\mathbb{s}}_{<j})$, making the skeleton a sequence of words. 
%The skeleton words $\hat{S}$ are predicted   %simply considered as subselection of content words in captions and they are
%autoregressively where each word is conditioned on the previously predicted skeleton word. 
The model is optimized with cross-entropy loss, trained using teacher forcing.
%This makes the skeleton a list of words.% conditioned based on left to right decoding. 
An attractive property is that the same architecture can be used to decode both the skeleton $\mathbb{S}$ and the caption $\mathbb{C}$.
%An attractive property in how we decompose the problem is that the same neural architecture can be used to learn to predict the simpler
%output space $\mathbb{S}$ instead of $\mathbb{C}$. 
 Moreover, the output tokens predicted in this stage are interpretable, and they are used to condition the second stage of our model. 
%\begin{myequ}
%\[
% $   \boldsymbol{\hat{\mathbb{s}}}^{\tau} \sim \prod_t Pr(\hat{\mathbb{s}}_k^t | \hat{\mathbb{s}}^{<t}, z_{\mathbb{I}})$
%\]
%\end{myequ}

\paragraph{Stage 2: Skeleton-based Caption Generation: } 
The second stage of training uses images and skeletons to generate captions $f_{\phi}: {\mathbb{I, S} \rightarrow \mathbb{C}}$. %information.
We experiment with 3 variants of conditioning predicted skeletons via encoding, decoding and autoencoding as shown in the model architecture in Fig.~\ref{fig:architecture}. The inputs, outputs and decoder attention conditioning for each stage are compared in Table \ref{tab:in_out_cond}. 
%We experiment with 3 variants of conditioning the %previously
%, as described here.

\begin{table}[]
\centering
\resizebox{0.49\textwidth}{!}{%
\begin{tabular}{l|ll|ll|l}
\hline \hline
 & \multicolumn{2}{l|}{\textbf{Stage 1}} & \multicolumn{2}{l|}{\textbf{Stage 2}} & \multirow{2}{*}{\textbf{Conditioning}} \\ \cline{1-5}
 & \textbf{Input} & \textbf{Output} & \textbf{Input} & \textbf{Output} &  \\ \hline
\textbf{SkeEnc} & $\mathbb{I}$ & $\mathbb{S'}$ & $\mathbb{I}$+$\mathbb{S'}$ & $\mathbb{C'}$ & $    \hat{\mathbb{c}}^{\tau} \sim \prod_t Pr(\hat{\mathbb{c}}^t | \hat{\mathbb{c}}^{<t}, g(z_{\mathbb{I}}, \hat{\mathbb{S}}))$
\\
\textbf{SkeAE} & $\mathbb{I}$ & $\mathbb{S'}$ & $\mathbb{I}+\mathbb{S'}$ & $\mathbb{S'}+\mathbb{C'}$ & $    \hat{\mathbb{c}}^{\tau} \sim \prod_t Pr(\hat{\mathbb{c}}_k^t | [\hat{\mathbb{{S}}} ; \hat{\mathbb{c}}^{<t} ], g(z_{\mathbb{I}}; \hat{\mathbb{S}}))$
\\
\textbf{SkeDec} & \multicolumn{2}{l|}{(no Stage 1)} & $\mathbb{I}$ & $\mathbb{S'}+\mathbb{C'}$ & $\hat{\mathbb{c}}^{\tau} \sim \prod_t Pr(\hat{\mathbb{c}}_k^t | [\hat{\mathbb{{S}}} ; \hat{\mathbb{c}}^{<t} ], z_{\mathbb{I}})$
\\
\hline \hline
\end{tabular}%
}
\caption{The inputs and outputs of the different models. In iterative refinement, $\mathbb{S'}$ is replaced by $\mathbb{C'}$.}
\label{tab:in_out_cond}
\end{table}

\paragraph{2a. SkeEncoding: } The predicted skeleton from the previous stage is used as input to the encoder. 
%This is demonstrated in the dotted box in the transformer encoder side in Figure \ref{fig:architecture}.
The image encoding and skeleton embeddings are fused with a unidirectional attention mechanism, called \textbf{text-as-side} (notated as $g$). In other words, we use the text representation as ``side information'' --- each transformed image feature unit can attend to other image feature units (self-attention) and text (cross-attention), but text cannot attend to image.
As shown in Fig.~\ref{fig:architecture}, this model has the dotted box in the Transformer encoder side, with the textual query, key, value ($Q_{w}$, $K_{w}$, $V_{w}$) and the visual counterpart attending to textual or visual key and value ($K_{v} + K_{w}$, $V_{v} + V_{w}$) with a visual query ($Q_{v}$).
We focus on the text-as-side attention mechanism as our preliminary results indicate that it leads to qualitatively better captions than image-text co-attention~\cite{lu2019vilbert}.

%We use separate Transformer encoders for image and text, and use the text representation as ``side information'' --- each transformed image feature unit can attend to other image feature units (self-attention) and text (cross-attention), but text cannot attend to image.
% CITE BEER's PAPER.  %and CITE co-attention.

\paragraph{2b. SkeDecoding: }
%This model predicts the skeleton first and then conditions the prediction of the caption in conjunction to the predicted skeleton within the same decoder autoregressively.
The skeleton and caption are concatenated and predicted by the same decoder.
This is not a two-staged model, as the model is trained to predict both skeleton and caption auto-regressively.
The model first predicts the skeleton words conditioned on the previously generated skeleton words, and then every token in the decoded caption attends to the entire predicted skeleton as well as the tokens of the caption decoded until that time step.
The dotted box in Transformer decoder of Fig.~\ref{fig:architecture} depicts this approach.
\useshortskip

\paragraph{2c. SkeAE: } To bring both the above models together, we simultaneously encode and decode the predicted skeleton. This brings the benefits of bidirectional attention on the input features (image and predicted skeleton words) and autoregressive attention on the re-predicted skeleton words while generating the caption.
In this case, both the dotted boxes on encoder and decoder sides in Fig.~\ref{fig:architecture} are active.
The encoding mechanism follows the $g$ function and the decoder prepends the caption generation task with the predicted skeleton. 
%The caption prediction model learns to copy the skeleton words predicted in the first stage (given as input) in the decoder output and autoregressively condition the caption on the predicted skeleton words.

%\paragraph{3. NoCaps $\mathbb{D_{x'y'}}$: } To demonstrate the effectiveness of the predicted skeleton to unseen domain scenarios, we use NoCaps or Novel Object Captioning dataset CITE. This data comprises of 4,500 validation and 10,600 test images. A popular large scale parallel training dataset, COCO comprises of 80 classes. The motivation behind the dataset is to differentiate the training conditions (typically, COCO), from the testing conditions using unseen Open Images CITE with about 600 classes. This inconsistency between distributions presents a gap between training and testing objectives that can be bridged by other annotations derived from ZZ CITE (Nocaps prior works). The participating criteria for this task are two fold. First, additional human annotated data is disallowed and second, ground truth object annotations from the Open Images dataset is forbidden. We satisfy both the requirements, as the CC data is noisy web crawled without any additional human interventions for annotations.

\section{Experiments and Results}

\paragraph{Hyperparameters: } Our transformer model uses 6 encoder and 6 decoder layers (unless specified otherwise), with 8 heads for multiheaded attention. 
%The 
Captions are subword-tokenized with a %vocabulary 
vocab size of 8,300. 
The models are optimized with Adam and an initial learning rate of $3.2e^{-5}$. %\ashish{Should this be 3.2e-6} This is num_cores*core_batch_size *10^(-6)*lr/256; (num_cores=32;batch_size=128) )
We use mini-batches of size 128, and train for 1M steps. The token embedding and filter sizes are both 512. We experimented with several values for both frequency thresholding for skeleton words at 20, 50, 100 and k at 2, 4, 8, 16 for top-k selection for multilabel classification model. We manually selected the values that optimize the model performance based on manual examination as conducting human evaluations with more hyperparameters is very expensive especially with unreliable automatic metrics.%, .hence we limited our study to this setting and focused on the concept of sub-selection.

%%%%%%%%%%%%%%%%%%%%%%%%%%%%%%%
\iffalse
\begin{table*}[]
\centering
\resizebox{0.7\textwidth}{!}{%
\begin{tabular}{l|l|l|l|l|l|l|l}
\hline \hline
\multirow{2}{*}{\textbf{Language}} & \multicolumn{4}{c|}{\textbf{Automatic Scores}} & \multicolumn{3}{c}{\textbf{Human Evaluations}} \\ \cline{2-8} 
 & \textbf{Baseline} & \textbf{SkeEncoding} & \textbf{SkeDecoding} & \textbf{SkeAE} & \textbf{Wins} & \textbf{Losses} & \textbf{Gains} \\ \hline
\textbf{French} & 0.9126 & 0.9020 & 0.8950 & 0.9033 & 34.43 & 29.53 & \textbf{+4.90} \\
\textbf{Italian} & 0.9010 & 0.8825 & 0.8605 & 0.8756 & 26.13 & 24.93 & \textbf{+1.20} \\
\textbf{German} & 0.7424 & 0.7256 & 0.7256 & 0.7310 & 35.23 & 33.93 & \textbf{+1.30} \\
\textbf{Spanish} & 0.9221 & 0.9108 & 0.8971 & 0.9141 & 34.03 & 34.33 & -0.3 \\
\textbf{Hindi} & 0.8511 & 0.8311 & 0.8240 & 0.8287 & 33.13 & 28.63 & \textbf{+4.50} \\ \hline \hline
\end{tabular}
}
%\vspace{-6pt}
\caption{CIDEr scores and human evaluation results for skeleton conditioned caption generation for multiple languages.}
%\vspace{-10pt}
\label{tab:multilingual}
\end{table*}
\fi
%%%%%%%%%%%%%%%%%%%%%%%%%%%%%%%

\subsection{Datasets}
\label{sec:datasets}
%The validity of our composable structure holds strict constraints on the coverage of $\mathbb{S}$ in the training data. 
%To satisfy this, we rely on a large scale, automatically curated dataset that is rich and diverse in semantic concepts. 
%To this end, we make use of Conceptual Captions .

% \noindent \textbf{1. Conceptual Captions (CC) $\mathbb{D_{xy}}$: } 
\paragraph{Conceptual Captions (CC): } 
%Free form natural language captions in the real world are more rich, diverse and noisy as compared to popularly used datasets such as MSCOCO \cite{DBLP:conf/eccv/LinMBHPRDZ14}.
CC \cite{DBLP:conf/acl/SoricutDSG18} is a large-scale dataset of 3.3M image-caption pairs covering a large variety of processed alt-texts from the web.
%In contrast to the popular MSCOCO dataset \cite{DBLP:conf/eccv/LinMBHPRDZ14}, it represents a step closer to having captioning annotations covering a wide variety of images and content present on the web with noisy yet diverse alt-text annotations. 
The focus of this work is on denoising noisy captioning datasets (web-scale, not human verified). 
Hence our experiments are focused on CC, which is a step closer to having large and diverse alt-texts from the web at the cost of being noisy.
In contrast, other popular datasets like COCO (size 120K) \cite{DBLP:conf/eccv/LinMBHPRDZ14} and Multi30k \cite{DBLP:conf/acl/ElliottFSS16}  are hand-annotated by humans and contain high quality images/captions. 
As a resource, CC is useful both for measuring progress on large-scale automatic captioning~\cite{DBLP:conf/acl/SoricutDSG18,changpinyo2019decoupled,alikhani2020cross,DBLP:conf/acl/ThapliyalS20}, as well as pre-training data for a variety of vision-and-language tasks~\cite{lu2019vilbert,chen19uniter,tan19lxmert,su20vlbert,li2020unicoder}.

\paragraph{Pre-processing: } CC might contain a long tail of spelling errors and other typos due to the automatic curation of the data. Therefore, we perform frequency based thresholding of the skeleton words to abate this noise. We experimented with several values for this hyperparameter and selected a minimum occurrence count as 50 that provides the desired balance between noise and vocabulary size. 
%, and selected this count as it optimizes the model performance.

%\noindent \textbf{2. Multilingual CC $\mathbb{D_{xy'}}$: } 
\paragraph{Multilingual CC: } 
%To demonstrate shifts in caption realizations, 
To demonstrate the cross lingual transferability of our  skeletons, we use automatic caption translations\footnote{We use the Google Cloud Translate API.} for CC, similar to the approach in~\cite{DBLP:conf/acl/ThapliyalS20}. 
%The target domain $\mathbb{D_{xy'}}$ has images from the same distribution as $\mathbb{D_{xy}}$, but the caption realization varies.
Note that the skeletons are learned from, and predicted in, English (not in the final target language), making English skeleton act as an \textit{interlingua}. 
Since multilingual captions are all pivoted on English skeletons, this nullifies the requirement to 
1) collect large-scale image-caption pairs in various language, and 
2) have access to linguistic tools to analyze captions in each language.
We perform experiments on 5 languages -- French, Italian, German, Spanish and Hindi -- which vary in word orders and token overlap with the English skeletons.

\paragraph{Conceptual Captions T2 test set:} 
For human evaluations across \emph{all languages}, we use %the
T2 test set used in the %CVPR'19 
Conceptual Captions Challenge\footnote{\url{http://www.conceptualcaptions.com/}}. 
It comprises of %1,000
1K out of domain images from the Open Images Dataset \cite{DBLP:journals/corr/abs-1811-00982}.%, and are out of domain for models trained on CC.
% Note that this same set of images are used for human evaluations across all the languages.

\subsection{Automatic Evaluation} 
\label{sec:automatic_metrics}

% Upto a maximum of 16 regions extracted from R-CNN are encoded. 
% For each skeleton form, we train both the classifier and generation based skeleton predictors, followed by skeleton-aware captioning model.
% For the first stage (skeleton prediction), we report precision, recall; for the second stage (caption generation), we report CIDEr scores and human evaluations. 

\paragraph{Skeleton Prediction:}
The goal of this stage is to extract key skeleton words from the image. 
%Hence 
We compute %the performance of each approach 
precision, recall and F-score as shown in Table~\ref{tab:clfvsgen}.
With the same %set of
labels (skeleton: nouns \& verbs), both classification and generation approaches have similar F-scores. 
However, precision is higher for generation and recall is higher for classification based predictions. Based on both qualitative observations and human %evaluations,
judgements, we note that generation approach was better, which shows that a higher precision is favorable in comparison to recall for this stage.
The %vocabulary
label size (of skeletons) in Table \ref{tab:clfvsgen} is approximately 5K.  

\begin{table}
\centering
\resizebox{0.48\textwidth}{!}{%
\begin{tabular}{l|c|c|c}
\hline \hline
 & \textbf{Iterative Refinement} & \textbf{Classification} & \textbf{Generation} \\ \hline
\textbf{Precision} & 35.75 & 23.22 & 36.66 \\ %\cline{2-5} 
\textbf{Recall} & 24.29 & 41.31 & 24.30 \\ %\cline{2-5} 
\textbf{F-score} & 28.92 & 29.73 & 29.23 \\ \hline \hline
\end{tabular}
}
\vspace{-6pt}
\caption{
{\small 
%Automatic metrics for various skeleton predictors. 
Performance of skeleton prediction stage. 
%Performance for both the stages are reported -- skeleton prediction (in Precision/Recall/F1), and skeleton-based captioning (using CIDEr).
Note that for classification and generation, the skeleton type used is `nouns \& verbs'.
}
}
%\vspace{-10pt}
\label{tab:clfvsgen}
\end{table}

\begin{table}[!t]
\centering
\resizebox{0.45\textwidth}{!}{%
\begin{tabular}{l|l|l|l}
\hline \hline
\textbf{Model} & \multicolumn{3}{l}{\textbf{CIDEr}} \\ 
\hline
\textbf{Baseline } (SOTA model) & \multicolumn{3}{l}{0.91} \cite{changpinyo2019decoupled} \\
\textbf{Impr. Img2Cap} & \multicolumn{3}{l}{1.00} \\ 
\textbf{Impr. Img2Cap (large)} & \multicolumn{3}{l}{0.99} \\ 
\hline \hline
\multirow{2}{*}{\begin{tabular}[c]{@{}l@{}} \textbf{Skeleton-based} \\ \end{tabular}} & \multicolumn{3}{c}{\textbf{Skeleton Type}} \\ 
%  & \begin{tabular}[c]{@{}l@{}}Nouns \& \\ Verbs\end{tabular} & Nouns & \begin{tabular}[c]{@{}l@{}}Sal Nouns \\ \& Verbs\end{tabular} \\ \hline
\cline{2-4}
 & \textbf{Nouns \& Verbs} & \textbf{Nouns only} & \textbf{Sal. Nouns \& Verbs} \\ \hline
~~~~~\textbf{SkeEncoding} & 0.99 & 0.97 & 0.94 \\ % \hline
~~~~~\textbf{SkeDecoding} & 0.99 & 0.99 & 0.96 \\ % \hline
~~~~~\textbf{SkeAE} & 0.99 & 0.96 & 0.94 \\ \hline \hline
\end{tabular}%
}
\vspace{-6pt}
\caption{
{\small 
Automatic metrics to compare various skeleton forms. % (such as Nouns \& verbs).
Img2Cap is the baseline (\textit{large} version refers to 12 encoder and decoder layers).
Note that these results use generation-based skeleton prediction.% was used for all these experiments.
}}
\vspace{-10pt}
\label{tab:results_quant}
\end{table}

%WRITE WHY WE PICKED CERRTAIN MODELS TO PERFORM HUMAN EVALUATION.

\noindent \textbf{Skeleton-based Caption Generation: }
We report multilingual IC performance of baseline and our dual-stage models using %the 
CIDEr %metric
in Table~\ref{tab:results_quant} (English) and Table~\ref{tab:multilingual} (multilingual).
Automatic metrics for captioning are based on 
%precision and recall of 
surface n-grams, and are not suitable to evaluate when the ground truth captions themselves are noisy.
In addition, we find that CIDEr is misleading \cite{alikhani2020cross, DBLP:conf/acl/SoricutDSG18, DBLP:conf/aaai/SeoSLHS20} and does not correlate with human evaluations %reported in 
(\S \ref{sec:human-evals}). 
All the 4 proposed skeleton variants are evaluated systematically for automatic metrics, as shown in the last column of Table \ref{tab:results_quant}. However, since the automatic scores are compared against a gold standard of noisy captions, they are not reliable. Hence we conducted manual evaluation to select the best performing skeleton variant. Out of the 4 skeleton variants, nouns and verbs performed better in  denoising and hence we demonstrated results for this variant for the remainder of the paper. We conducted further experimentation on nouns and verbs on the three models of dual staged captioning, controllability and cross-lingual transferability. 
%Upon manual examination of 100 examples, using nouns and verbs as skeletons performs better in comparison to the rest of the skeletons. Hence the downstream experimentations in the rest of the paper are conducted with nouns and verbs as the skeleton form.

%, and hence share the same problems as described above for skeleton generation task.

\begin{figure*}[t!]
\centering
\includegraphics[trim=0cm 8cm 0cm 1.5cm,width=0.98\linewidth]{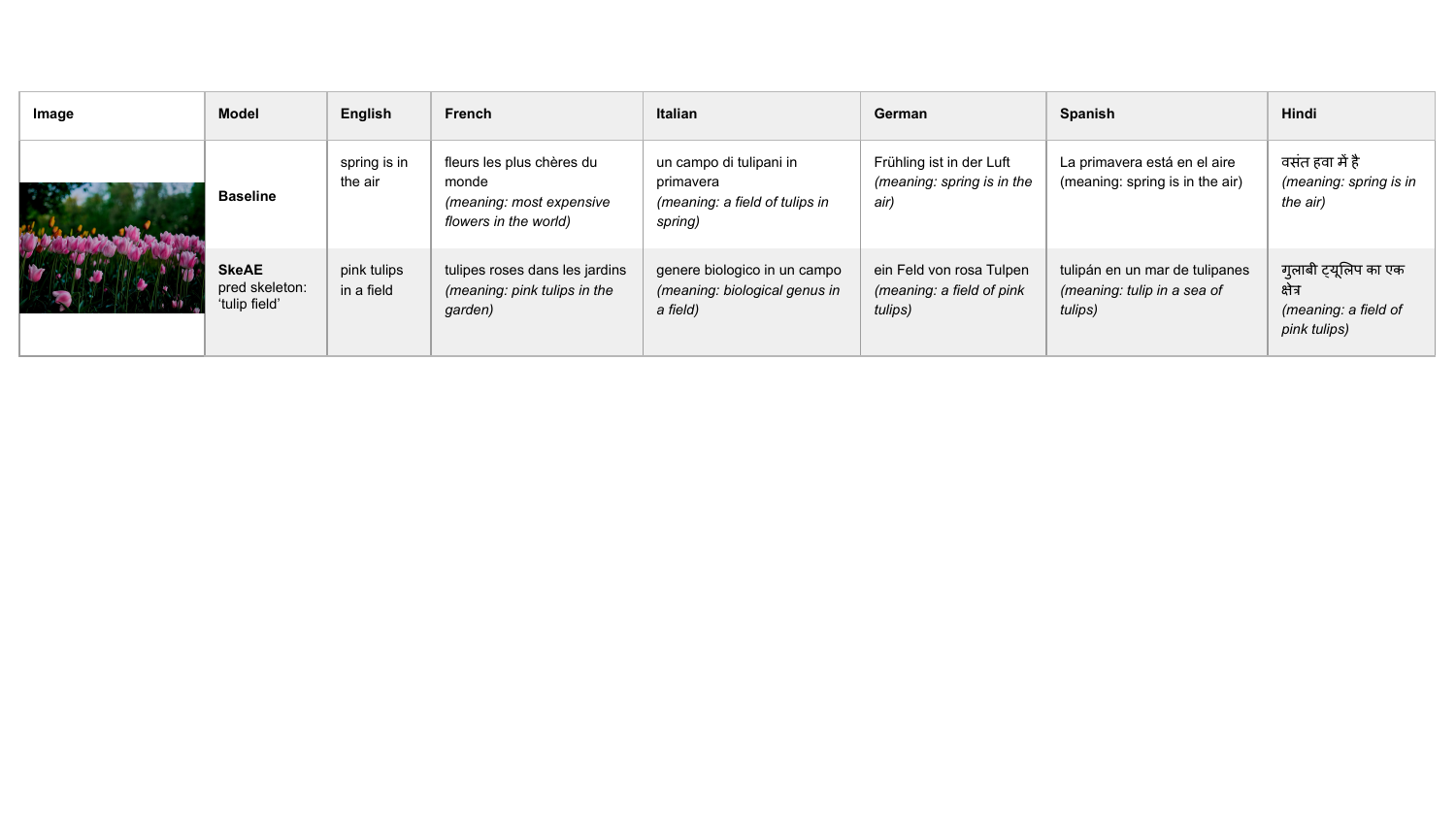}
\caption{{\small Captions generated by baseline and our dual staged approach in 6 languages and their corresponding translations.}}
%\vspace{-6pt}
\label{fig:multilingual}
%\vspace{-15pt}
\end{figure*}

\paragraph{Multilingual captioning: }Note that the skeletons are always in English, trained using annotations over the original English CC dataset.
Cross-lingual results on val data of Multilingual CC are presented in Table \ref{tab:multilingual}.
In addition to the data noisiness, a reason for slightly lower performance for non-English captions is probably noisy translation artifacts. For example, corresponding 
caption in the Hindi dataset 
for English caption `She is gazing at the \textit{fall colors}' is `{\dn vh Egrt\? r\2go\2 kF aor d\?K rhF h\4}'
%\begin{hindi}
 %{\hindifont
% वह गिरते रंगों की ओर देख रही है।
 %}
%\end{hindi}
(translation: She is looking at the \textit{falling colors}.)
Translation errors (such as `fall' colors to `falling' colors) introduce noise in the non-English datasets.
Figure~\ref{fig:multilingual} presents an example of output multilingual captions for the baseline and our SkeAE approach.

\paragraph{Unpaired Image Captioning: }
A natural extension to our approach is for the caption generator to rely purely on %the generated
predicted skeleton, and not use image features.
This is a harder problem, but eliminates altogether, the need for image-caption pairs because the second stage (skeleton to caption) can be trained on a large text-only corpus.
% The first stage predicts the skeleton from the image, and the second stage generates a caption just from the predicted skeleton (without relying on the image). 
%If the predicted skeleton is adequate to compose a caption from, the role of image features can be minimized in the second stage of training. 
In this direction, within the scope of CC dataset, we investigate 1) with and without using image features in the second stage, 2) using ground truth skeleton (GTSke) to get an estimate of the upper bound on unpaired captioning 3) comparing the upper bound to the predicted skeleton (PredSke). These results are presented in Table \ref{tab:unpaired}.
When image features are ignored, CIDEr drops by only 8  points when only predicted skeletons are used for caption generation compared to the baseline. This initial result shows that skeletons are a promising direction towards unpaired captioning.
%in order to get an estimate of the upper bound, we used ground truth skeleton (GTSke) and compared it with predicted skeleton (PredSke)
%using ground truth skeleton (GTSke) and comparing it with predicted skeleton (PredSke) to get an estimate of the upper bound

\subsection{Human Evaluations}
\label{sec:human-evals}

\begin{table}[!t]
\centering
\resizebox{0.48\textwidth}{!}{%
\begin{tabular}{l|l|l|l|l}
\hline \hline
\textbf{Language} &\textbf{Baseline} & \textbf{SkeEncoding} & \textbf{SkeDecoding} & \textbf{SkeAE} \\ \hline
\textbf{French} & 0.91 & 0.90 & 0.89 & 0.90  \\
\textbf{Italian} & 0.90 & 0.88 & 0.86 & 0.87  \\
\textbf{German} & 0.74 & 0.72 & 0.72 & 0.73  \\
\textbf{Spanish} & 0.92 & 0.91 & 0.89 & 0.91 \\
\textbf{Hindi} & 0.85 & 0.83 & 0.82 & 0.82  \\ \hline \hline
\end{tabular}
}
%\vspace{-6pt}
\caption{ {\small CIDEr scores for skeleton (form: Nouns \& Verbs, prediction approach: generation) conditioned caption generation for multiple languages.}}
\vspace{-10pt}
\label{tab:multilingual}
\end{table}

\begin{table}[!t]
\centering
\small
\begin{tabular}{l|l}
\hline \hline
\textbf{Model Enc Input} & \textbf{CIDEr} \\ \hline
PredSke + Img (Paired) & 0.99 \\ 
PredSke (Unpaired) & 0.91 \\ \hline
GtSke + Img (Paired Headroom) & 4.62 \\
GtSke (Unpaired Headroom) & 4.48 \\ \hline
 \hline 
\end{tabular}%
%\caption{CIDEr scores on val data with ablations for unpaired captioning.}
\vspace{-6pt}
\caption{\small{Ablations on val data for unpaired captioning.}}
\vspace{-10pt}
\label{tab:unpaired}
\end{table}

\begin{figure}[t!]
\centering
\includegraphics[trim=3cm 2cm 1cm 1.5cm,width=0.95\linewidth]{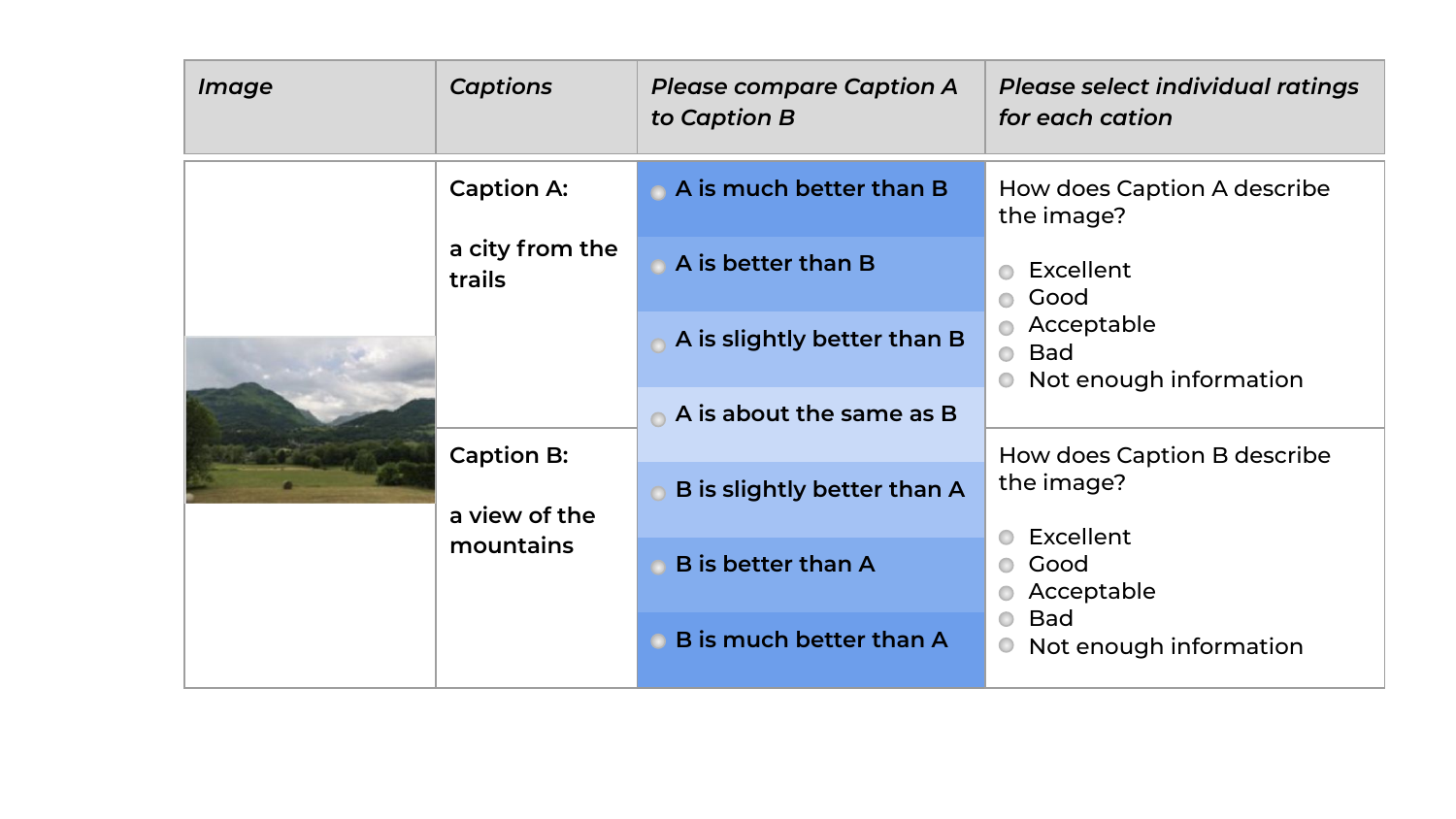}
\vspace{-6pt}
\caption{{\small Human evaluation interface: We ask raters to: 
1) compare the two captions (relative), 2) give ratings for each caption (absolute). Human annotators are asked to indicate the better caption relevant to the image. 
% We use only relative ratings for model comparisons and  absolute ratings only to cross-check.
}}
\vspace{-10pt}
\label{fig:humanevalinterface}
\end{figure}

Automatic metrics often have been found not to correlate well with human scores~\cite{DBLP:conf/eacl/KilickayaEIE17, alikhani2020cross} and do not fare well when ground truth text is noisy. So we conduct extensive human evaluations where captions for each image are evaluated both in relative preferences and absolute scale \cite{DBLP:conf/acl/ThapliyalS20}.
As mentioned above, we use the T2 test set of 1000 images, each rated by 3 distinct annotators.
The interface of this evaluation is displayed in Figure \ref{fig:humanevalinterface}.
While comparing two models side-by-side, they are randomly assigned `A' or `B' in the interface for each image to avoid any rater bias.
%The ratings gathered include both relative and absolute.

\begin{table}[t!]
\centering
\resizebox{0.45\textwidth}{!}{%
\begin{tabular}{l|l|l|l|l}
\hline \hline
\textbf{Approach} & \textbf{Skeleton}    & \textbf{Wins} &\textbf{Losses} & \textbf{Gains} \\ \hline
\textbf{SkeEncoding}       & Nouns \& Verbs &  39.34 & 28.33 & {\bf +11.0} \\ 
\textbf{SkeAE}             & Nouns \& Verbs &  39.34 & 32.63 & +6.7 \\
\textbf{SkeDecoding}       & Nouns \& Verbs &  34.83 & 34.53 & +0.3 \\
\textbf{SkeEncoding}       & Iterative Refinement & 19.62 & 20.52 & -1.1 \\ \hline \hline
\end{tabular}%
}
%\vspace{-6pt}
\caption{\small{Human evaluation scores of different approaches and skeletons on English (vs the Img2Cap baseline).}}
%\vspace{-10pt}
\label{tab:he_eng}
\end{table}

\begin{table}[t!]
\centering
\resizebox{0.28\textwidth}{!}{%
\begin{tabular}{l|l|l|l}
\hline \hline
\textbf{Language} & \textbf{Wins} & \textbf{Losses} & \textbf{Gains} \\ \hline
\textbf{French} & 31.43 & 29.53 & \textbf{+1.9} \\
\textbf{Italian} & 26.13 & 24.93 & \textbf{+1.2} \\
\textbf{German} & 35.23 & 33.93 & \textbf{+1.3} \\
\textbf{Spanish} & 34.03 & 34.33 & -0.3 \\
\textbf{Hindi} & 33.13 & 28.63 & \textbf{+4.5} \\ \hline \hline
\end{tabular}
}
%\vspace{-6pt}
\caption{\small{Human evaluation results for skeleton (form: nouns \& verbs, prediction approach: generation with SKeEnc) conditioned caption generation for multiple languages.}}
%\vspace{-10pt}
\label{tab:he_multilingual}
\end{table}

\paragraph{Relative Ratings:} 
For each image we ask the raters to choose the most relevant caption. % that is most relevant to the image. 
Comparing Caption A to Caption B, raters can select relative options as shown in the third column in Figure \ref{fig:humanevalinterface}. \textit{Wins} are the percentage of images where at least 2 out of 3 annotators voted for caption generated with our approach. \textit{Losses} are percentage of images where at least 2 out of 3 annotators voted for caption generated with Img2Cap approach. We compute \textit{gains} in this side by side relative evaluation as $Gains_{relative}$ = Wins - Losses.

\paragraph{Absolute Ratings:}
We also gather absolute rating for each of the 2 captions per image. Each caption is rated as acceptable if at least 2 out of 3 annotators rate it as %one among
\textit{acceptable}, \textit{good} or \textit{excellent}. %In this way, absolute rating is computed as:\\
\textit{$Gains_{absolute} = Accept_{our\_approach} - Accept_{baseline}$}. %However they are not used in this quantitative analysis.
We use them only to validate the ratings such that, for example, an ``Excellent'' rated caption is not annotated as inferior to a ``Bad'' rated caption for the same image. These ratings are collected to double check the results of the relative rating as well.

These scores are presented in Table \ref{tab:he_absolute}. The top part of the table indicate the absolute ratings in terms of Good and OK performance for multilingual captions. The second part of the table show the same scores when baseline model is compared with the corresponding model and skeleton combination. Each model i.e baseline and the proposed model in each row are rated individually (not relative to one another). The last two columns indicate the performance shift of the corresponding proposed model with respect to the baseline in each of the Good and OK categories.

\begin{table*}[h]
\centering
\resizebox{0.85\textwidth}{!}{%
\begin{tabular}{l|l|l|l|l|l|l|l}
\hline \hline
\textit{Row no.} & \textbf{Language} & \textbf{Good Baseline} & \textbf{Good SkeAE} & \textbf{OK Baseline} & \textbf{OK SkeAE} & \textbf{Gains in Good} & \textbf{Gains in OK}\\ \hline
1 & \textbf{French} & 34.63 & 35.04 & 61.36 & 60.66 &  +0.40 & -0.70 \\
2 & \textbf{Italian} & 35.14 & 35.44 & 60.86 & 62.56 & +0.30 & +1.70 \\
3 & \textbf{German} & 43.64 & 41.04 & 67.27 & 68.07 &  -2.60 & 0.80 \\
4 & \textbf{Spanish} & 48.15 & 46.55 & 74.37 & 74.67 &  -1.60 & +0.30 \\
5 & \textbf{Hindi} & 59.96 & 66.17 & 85.99 & 87.99 & +6.21 & +2.00 \\
\hline
\textit{Row no.} & \textbf{Model} & \textbf{Good Baseline} & \textbf{Good Model} & \textbf{OK Baseline} & \textbf{OK Model} & \textbf{Gains in Good} & \textbf{Gains in OK}\\ \hline
6 & \textbf{Unpaired} &  57.36 & 55.06 & 86.48 & 84.28 &  -2.30 & -2.20 \\
7 & \textbf{SkeEnc (Iterative Refinement)} &  63.76 & 62.36  & 87.89 & 87.49 & -1.40 & -0.40 \\
8 & \textbf{Nouns and Verbs (SkeEnc)} & 66.47 & 63.66 & 89.39 & 88.89 & +2.81 & +0.50 \\
9 & \textbf{Nouns and Verbs (SkeAE)} &  51.55 & 56.66 & 79.68 & 83.18 & + 5.01 & +3.40  \\
\hline \hline
\end{tabular}
}
%\vspace{-6pt}
\caption{\small{Absolute ratings in percentages in Human Evaluations.}}
%\vspace{-10pt}
\label{tab:he_absolute}
\end{table*}

\paragraph{Results:} 
Table \ref{tab:he_eng} presents the human ratings for English captions using different skeletons. From this, we observe the following:

\noindent \textbf{{\textit{(a) Dual Staging helps: }}} Our dual staged models with skeletons (SkeEnc, SkeDec, SkeAE) show gains compared to the improved baseline Img2Cap model. Most notably, it shows that the `Nouns \& Verbs' skeletons significantly improves SkeEncoding model attaining the most significant gain, followed by SkeAE and then SkeDecoding. 

\noindent \textbf{{\textit{(b) Subselecting content words helps: }}} Using the same dual staged SkeEnc model without subselecting content words in the form of iterative refinement does not show any improvement in performance, supporting the hypothesis that sub-selecting content skeleton from noisy captions improves the overall caption quality.

%Most notably, it shows that the \textit{Nouns \& Verbs skeletons significantly improve} over the end-to-end Img2Cap baseline, with the 
%SkeEncoding model attaining the most significant gain, followed by SkeAE and then SkeDecoding.
%The dual-stage approach does not show improvement using iterative refinement, %over the baseline 
%supporting the hypothesis that selecting content skeleton improves the overall caption quality.
% \beer{we should say that the process of extracting nouns and verbs are important.}
\noindent \textbf{{\textit{(c) Cross-lingual skeleton transfer: }}} Table \ref{tab:he_multilingual} presents our human evaluation scores for captions in other target languages.
We observe gains from the skeleton-based approach for 4 out of 5 languages, and only a slight loss for the fifth language i.e., Spanish, showing the effectiveness of cross-lingual transferability of the skeleton words. Our speculation for this is probably due to the dialect differences. The translation model that we used for Spanish is a mix of Spain Spanish and Latin American Spanish, with Latin American Spanish dominating. The evaluation was done by raters from Spain. The dialects are sufficiently different that it would impact the absolute scores. 

\begin{figure*}[tbh]
\centering
\includegraphics[trim=0cm 8.8cm 0cm 0cm,width=0.8\linewidth]{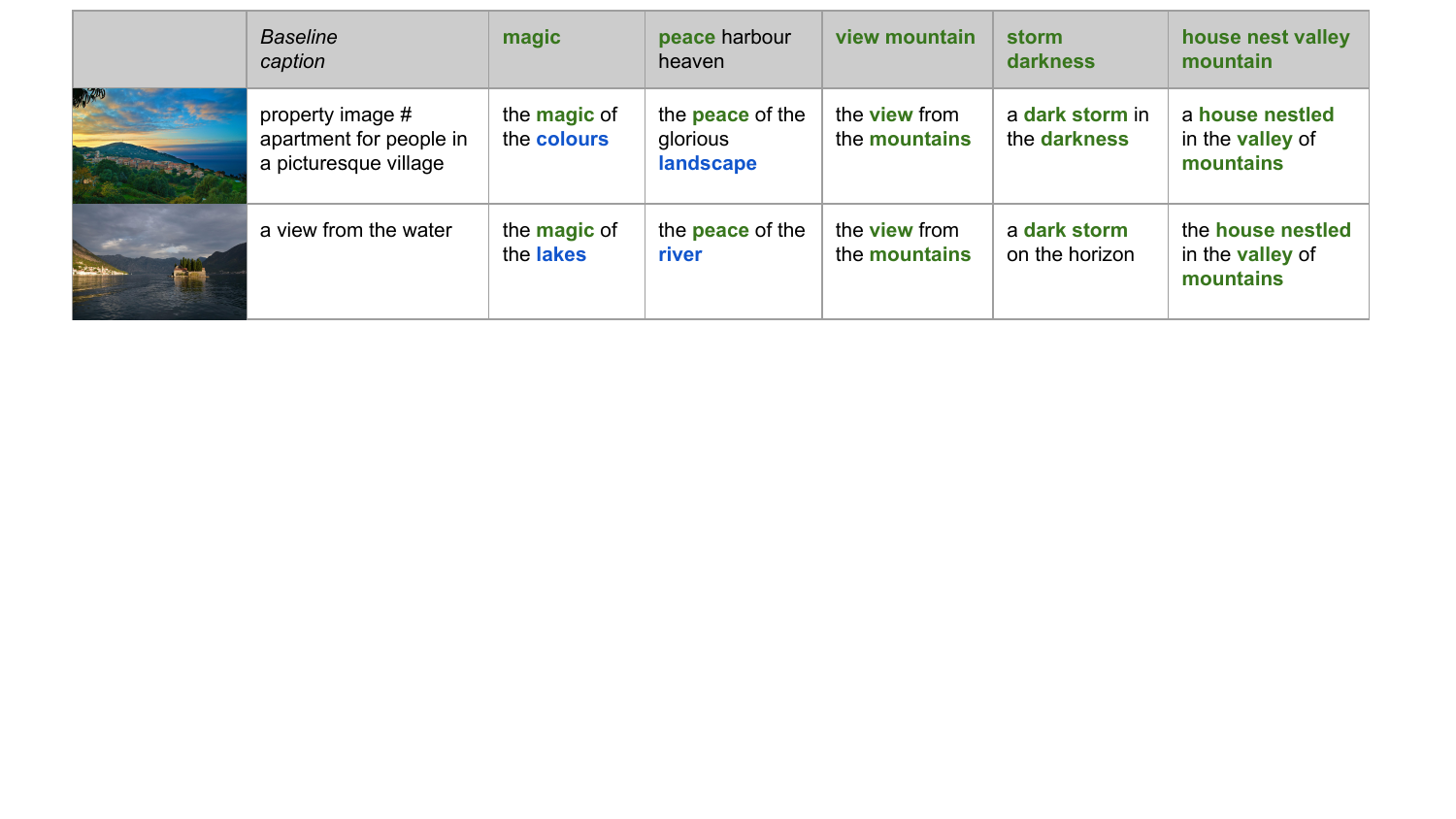}
%\vspace{-6pt}
\caption{{\small Controllability: Effect of guiding the information through skeleton.  As observed, the caption incorporates information from the skeleton that is consistent with the image. For example, in the second column of the top row, we see that peace is incorporated while harbor and heaven are not. The relevant skeleton words in other columns guide the captions accordingly.}}
\vspace{-10pt}
\label{fig:alter_qual}
\end{figure*}

\begin{figure}[t!]
\centering
\includegraphics[trim=0cm 0cm 0cm 0cm,width=0.65\linewidth]{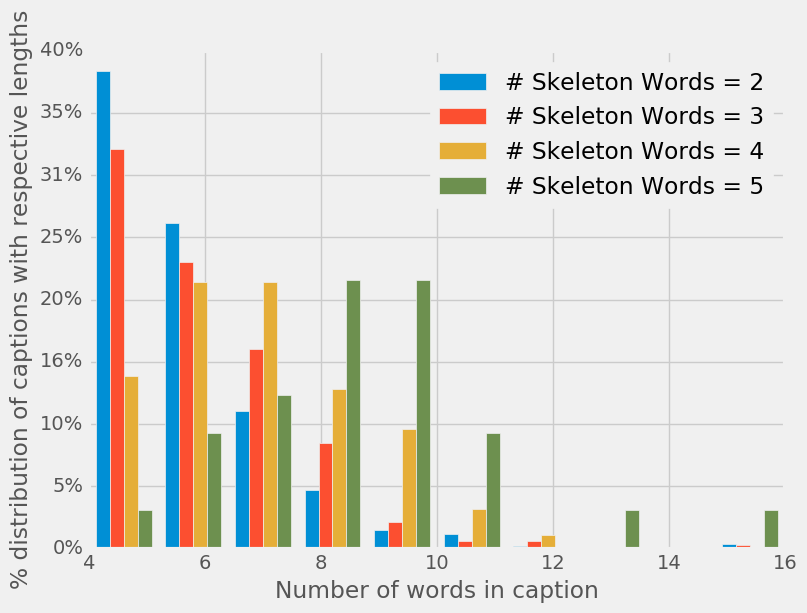}
%\vspace{-6pt}
\caption{ { \small
Quantitative relationship between the number of skeleton words and caption length.} 
}
%\vspace{-15pt}
\label{fig:length_chart}
\end{figure}

\subsection{Cross-modal Discourse Coherence} 
\label{sec:discourse}
To understand where the improvements quantified in Table~\ref{tab:he_eng} come from, we turn to the notion of discourse coherence.
\citet{alikhani2020cross} introduce %the notion of 
multimodal discourse coherence relationships between image-caption pairs.
%For instance, 
For instance, a caption describing visually recognizable aspects of the image, such as `people' or `cake', is annotated using a \emph{Visible} relation; in contrast,
a \emph{Meta} relation corresponds to a caption containing details regarding how/when/where the image was captured, such as in `warm summer afternoon', while
a \emph{Story} relation implies that the caption describes some potentially non-visible context behind the scene depicted in the image, such as `fifth anniversary'.
%For the full list of labels, please refer to the paper.

\begin{table}[t!]
\centering
\resizebox{0.4\textwidth}{!}{%
\begin{tabular}{l|c|c|c|c}
    \hline \hline
        & \multicolumn{3}{c|}{\textbf{Counts}} & \multirow{2}{*}{\textbf{Human Evals}} \\
        \cline{2-4}
        & \textbf{Baseline} & \textbf{Ours} & \textbf{Change} & \\
    \hline
    \emph{Visible} & 605 & 640 & +5.79\% & +10.93\%\\
    \emph{Meta} & 245 & 226 & -7.76\% & +13.06\% \\
    \emph{Story} & 129 & 108 & -16.28\% & +10.08\% \\
    \hline
\end{tabular}
}
%\vspace{-6pt}
\caption{ \small{
Analysis of multimodal discourse coherence relations for baseline and our model on T2 dataset.
The last column shows the relative human evaluation gains over baseline caption of each type.
Other relations with small counts are ignored in the above analysis.}
}
\vspace{-10pt}
\label{tab:coherence}
\end{table}

We hypothesize that our multi-stage approach of skeleton-based IC results in the generation of more captions of \emph{Visible} type, as 
%a result of 
the intermediate skeleton predictor is trained to predict nouns and verbs from the image.
%A caption conditioned on such a skeleton is more likely to describe the visual content of an image, and, as a result, produce captions that are in a \emph{Visible} relation with the image.
As observed in \S \ref{sec:human-evals}, as SkeAE model performs better compared to the SkeEncoding and SkeDecoding models, we analyze the downstream captions based on SkeAE architecture. 
To assess this effect, we train %and deploy 
the relation classifier described in Sec.~4 of \cite{alikhani2020cross}, and obtain discourse relation labels for captions generated on T2-test images, by both the baseline Img2Cap and our SkeAE  models. 
Table~\ref{tab:coherence} (Counts columns) quantifies the shift of relation label distribution towards the \emph{Visible} coherence relation, confirming our hypothesis. 
%Additionally, 
We also study the breakdown by coherence relations using the results from our human evaluations on the English captions.
Table~\ref{tab:coherence} (Human Evals column) reports this breakdown, indicating that,
of the 11.01\% gains on human evals from Table~\ref{tab:he_eng}, the shift from non-Visible to Visible discourse captions is associated with clear increases in preference from the human raters.
This is attributable to the fact that human raters are more likely to prefer captions that are in a \emph{Visible} relation with 
%respect to 
the image, and therefore the shift towards generating \emph{Visible}-type captions can be positively quantified in terms of human preference.

%067856ad3c5668c0: sunglasses
% qualitative example: 00c2db0fd32a0f2b
% 0287310d61949134: cig example

\section{Controllability: Qualitative Discussion }

%The \% of longer captions increase for skeleton of longer length.

\begin{figure}[t!]
\centering
\includegraphics[trim=0cm 0cm 0cm 0cm,width=0.96\linewidth]{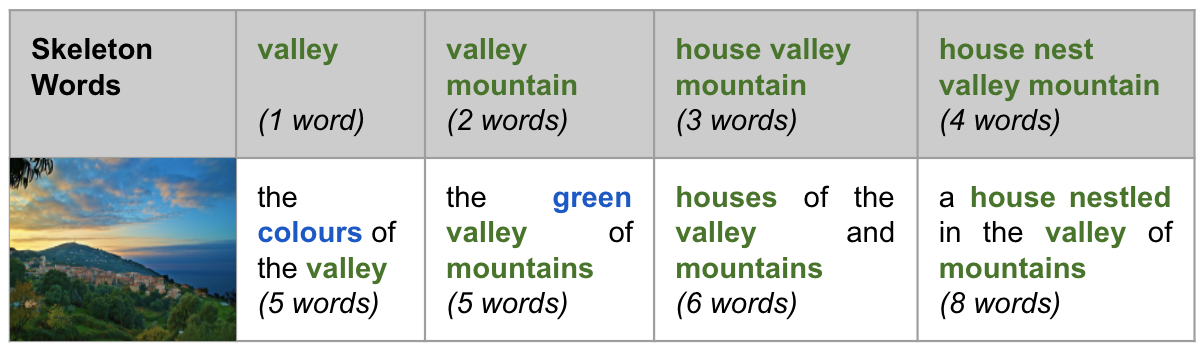}
\vspace{-6pt}
\caption{ {\small Controllability: Effect of varying the number of words in the skeleton on the generated caption length.}
%We see that the generated caption gets longer and more informative as it incorporates an increasing number of skeleton words.
}
\vspace{-15pt}
\label{fig:length_qual}
\end{figure}

The dual-stage modeling using skeelton decomposition can be a double-edged sword: 
it can be an information bottleneck, limiting the ability to train the model in an end-to-end manner; but, it brings forth the advantage of increased interpretability and thereby the ability to use the intermediate stage results to control the %control or guide the 
final caption. 
We present aspects of caption controllability by altering the skeleton to explore effects on caption length, informativeness, and gender specificity. % Note that this analysis only shows that our approach guides the caption based on the %variation in skeleton.
This section discusses the utility of this dual staged model for controllability qualitatively with SkeAE architecture. 
Automatic intervention at the skeleton level involves non-trivially selecting related concepts for each image, and we leave this for follow-up work. 
Instead, we present an empirical study only to semi-automatically control gender specificity in two of the languages.
We plan to conduct experiments to compare with other models \cite{DBLP:conf/cvpr/ZhengLW19, DBLP:conf/cvpr/ChenJWW20} focused on controllability 
%and automatically selecting different but relevant skeleton words in the future work.
 for follow-up work.

\noindent \textbf{Effect of length of skeletons on captions:} For applications that %impose 
limit the caption lengths due to UI restrictions, the ability to control the length is important. 
The length of the skeleton correlates with the number of caption words, as shown in Figure \ref{fig:length_chart}. 
For 2 or 3 skeleton words, the percentage of captions monotonically decreases with the number of caption words, with the mode at 4-word captions. Thus, for skeletons of size 2, captions of length 4 are much more frequent than captions of length 6 or 8.
% Note: let's use "skeletons" for plural (I changed it throughout the paper).
For longer skeletons, we see that the mode shifts to the right: with skeletons of size 5, the caption length peaks between 8 and 10 words.  %Thus captions of length 4 are less frequent than captions of length 8 or 10 and so on.%, which in turn are more frequent than those with size twelve, fourteen or sixteen.
%Note that, 
%The third and the fourth bar (i.e with 4 and 5 number of skeleton words) have an inverted cup pattern.
%They increase upto a point and then begin to decrease. This inversion point is at a higher caption length when the number of skeleton words is 5 compared to when it is 4. 
% The length can also be proxied to informativeness of the skeleton. 
%The effect of varying the skeleton size on the caption length is qualitatively shown in Figure 
Fig \ref{fig:length_qual} illustrates this qualitatively.

\noindent \textbf{Effect on gender specificity: }
%As a preliminary observation, 
%We note that 
%The current models are quite clumsy in generating gender-debiased captions.%, and are prone to make embarrassing errors.
Current models often make embarrassing mistakes when generating captions that mention gender.
The availability of a skeleton provides %allows us to have 
a direct handle for human-in-the-loop correction of such biases, at a pre-caption-generation stage.
This is more robust compared to caption post-processing, especially for highly inflected languages.
To illustrate this, %control ability,
we compare the number of times `man' appears in the captions generated by our baseline versus our dual-stage model after automatically modifying the skeleton (replacing `man' to the gender-neutral word `person' in the skeleton).
Over the T2 dataset, the baseline caption generates `man' 13 times, and the automatic control mechanism via our model
%for the dual-staged model 
reduces this by 46\% (to 7 occurrences) in English. In Hindi, the equivalent of `man' ({\dn aAdmF})
%({\hindifont आदमी}) 
is generated 10 times, and it is reduced to a gender neutral word
({\dn \326wyE\3C4w})
by 70\% (to 3 occurrences).
%{\hindifont व्यक्ति}) by 70\% (to 3 occurrences).

%(`आदमी')

% Man 13 -> 7 (English) 
% Woman 7 -> 7 (English)
% Person 63 -> 71 (English)

% and 10->3 in Hindi
% NOT SURE ABOUT THIS: WANT TO ADD AN EXAMPLE.

\noindent \textbf{Effect of guiding information through skeleton: }
The skeleton acts as a knob enabling the model to describe different attributes of the image in the caption. 
Figure~\ref{fig:alter_qual} presents an example of how varying the skeletons for two different images affect their captions.
The words highlighted in green are derived from the skeleton and  %words and the %words highlighted 
the ones highlighted in blue are image-related words.

\section{Conclusions}

%Most image captioning models rely on clean image-caption pairs limiting their ability to scale to large, noisy and uncurated  data available on web.
Scaling image captioning models practically mandates training on noisy and uncurated data available on web.
%the richness in captions.
Our work presents an approach that denoises learning from such large yet diverse web-scaled data with alt-text annotations by sub-selecting content as intermediate skeletons.
We experimentally demonstrate that this approach 
%does not change CIDEr much, it 
improves the captions significantly %when trained on Conceptual Captions 
in human evaluations on out-of-domain test data by converting meta and story like captions to more visually   informative captions. 
We also demonstrate the transferability of English skeleton words to improving captions in five other languages.
Additionally, the natural-language interpretable skeleton layer gives us a way to better control and perform human-in-the-loop corrections of model predictions. 
%We demonstrate analyses on controlling the length, informativeness and gender specificity of captions. 
We believe that this is a promising direction towards unpaired IC and also has potential for semi-automatic interventions to correct or interact with the skeletons to guide the final captions. 

In this work, our main focus is denoising alt-text captions using skeletons and using them for cross-lingual captioning. In future, we plan to explore the effect of denoising in pretraining large multimodal models (BLIP \cite{DBLP:journals/corr/abs-2201-12086}, UNITER \cite{DBLP:conf/eccv/ChenLYK0G0020}, ViLBERT \cite{DBLP:conf/nips/LuBPL19}) as base architectures by automatically cleaning captions, similar to how BLIP has an additional classifier to subselect captions that are not noisy.
% that eliminates the need for expensive image-caption pairs for training.
\textit{Appendix \ref{sec:broader_impact} presents a broader impact of our work.}

% Entries for the entire Anthology, followed by custom entries

%As a result, we are discouraging the model to learn noisy correlations and encouraging it to learn causality between content based skeleton words and captions.

%\section*{Acknowledgements}

% Entries for the entire Anthology, followed by custom entries
\bibliography{anthology,custom}

\begin{thebibliography}{75}
\expandafter\ifx\csname natexlab\endcsname\relax\def\natexlab#1{#1}\fi

\bibitem[{Alikhani et~al.(2020)Alikhani, Sharma, Li, Soricut, and
  Stone}]{alikhani2020cross}
Malihe Alikhani, Piyush Sharma, Shengjie Li, Radu Soricut, and Matthew Stone.
  2020.
\newblock Cross-modal coherence modeling for caption generation.
\newblock In \emph{Proceedings of the 58th Annual Meeting of the Association
  for Computational Linguistics}, pages 6525--6535.

\bibitem[{Anderson et~al.(2018)Anderson, He, Buehler, Teney, Johnson, Gould,
  and Zhang}]{anderson2018bottomup}
Peter Anderson, Xiaodong He, Chris Buehler, Damien Teney, Mark Johnson, Stephen
  Gould, and Lei Zhang. 2018.
\newblock Bottom-up and top-down attention for image captioning and visual
  question answering.
\newblock In \emph{CVPR}.

\bibitem[{Ba et~al.(2016)Ba, Kiros, and Hinton}]{ba2016layer}
Jimmy~Lei Ba, Jamie~Ryan Kiros, and Geoffrey~E. Hinton. 2016.
\newblock Layer normalization.
\newblock \emph{arXiv preprint arXiv:1607.06450}.

\bibitem[{Barrault et~al.(2018)Barrault, Bougares, Specia, Lala, Elliott, and
  Frank}]{DBLP:conf/wmt/BarraultBSLEF18}
Lo{\"{\i}}c Barrault, Fethi Bougares, Lucia Specia, Chiraag Lala, Desmond
  Elliott, and Stella Frank. 2018.
\newblock \href {https://doi.org/10.18653/v1/w18-6402} {Findings of the third
  shared task on multimodal machine translation}.
\newblock In \emph{Proceedings of the Third Conference on Machine Translation:
  Shared Task Papers, {WMT} 2018, Belgium, Brussels, October 31 - November 1,
  2018}, pages 304--323. Association for Computational Linguistics.

\bibitem[{Changpinyo et~al.(2019)Changpinyo, Pang, Sharma, and
  Soricut}]{changpinyo2019decoupled}
Soravit Changpinyo, Bo~Pang, Piyush Sharma, and Radu Soricut. 2019.
\newblock Decoupled box proposal and featurization with ultrafine-grained
  semantic labels improve image captioning and visual question answering.
\newblock In \emph{EMNLP-IJCNLP}.

\bibitem[{Chen et~al.(2020{\natexlab{a}})Chen, Jin, Wang, and
  Wu}]{DBLP:journals/corr/abs-2003-00387}
Shizhe Chen, Qin Jin, Peng Wang, and Qi~Wu. 2020{\natexlab{a}}.
\newblock \href {http://arxiv.org/abs/2003.00387} {Say as you wish:
  Fine-grained control of image caption generation with abstract scene graphs}.
\newblock \emph{CoRR}, abs/2003.00387.

\bibitem[{Chen et~al.(2020{\natexlab{b}})Chen, Jin, Wang, and
  Wu}]{DBLP:conf/cvpr/ChenJWW20}
Shizhe Chen, Qin Jin, Peng Wang, and Qi~Wu. 2020{\natexlab{b}}.
\newblock \href {https://doi.org/10.1109/CVPR42600.2020.00998} {Say as you
  wish: Fine-grained control of image caption generation with abstract scene
  graphs}.
\newblock In \emph{2020 {IEEE/CVF} Conference on Computer Vision and Pattern
  Recognition, {CVPR} 2020, Seattle, WA, USA, June 13-19, 2020}, pages
  9959--9968. {IEEE}.

\bibitem[{Chen et~al.(2020{\natexlab{c}})Chen, Li, Yu, Kholy, Ahmed, Gan,
  Cheng, and Liu}]{chen19uniter}
Yen-Chun Chen, Linjie Li, Licheng Yu, Ahmed~El Kholy, Faisal Ahmed, Zhe Gan,
  Yu~Cheng, and Jingjing Liu. 2020{\natexlab{c}}.
\newblock {UNITER}: {Learning UNiversal Image-TExt Representations}.
\newblock In \emph{ECCV}.

\bibitem[{Chen et~al.(2020{\natexlab{d}})Chen, Li, Yu, Kholy, Ahmed, Gan,
  Cheng, and Liu}]{DBLP:conf/eccv/ChenLYK0G0020}
Yen{-}Chun Chen, Linjie Li, Licheng Yu, Ahmed~El Kholy, Faisal Ahmed, Zhe Gan,
  Yu~Cheng, and Jingjing Liu. 2020{\natexlab{d}}.
\newblock \href {https://doi.org/10.1007/978-3-030-58577-8\_7} {{UNITER:}
  universal image-text representation learning}.
\newblock In \emph{Computer Vision - {ECCV} 2020 - 16th European Conference,
  Glasgow, UK, August 23-28, 2020, Proceedings, Part {XXX}}, volume 12375 of
  \emph{Lecture Notes in Computer Science}, pages 104--120. Springer.

\bibitem[{Cornia et~al.(2019)Cornia, Baraldi, and
  Cucchiara}]{DBLP:conf/cvpr/CorniaBC19}
Marcella Cornia, Lorenzo Baraldi, and Rita Cucchiara. 2019.
\newblock Show, control and tell: A framework for generating controllable and
  grounded captions.
\newblock In \emph{Proceedings of the IEEE Conference on Computer Vision and
  Pattern Recognition}, pages 8307--8316.

\bibitem[{Cornia et~al.(2020)Cornia, Stefanini, Baraldi, and
  Cucchiara}]{cornia20m2}
Marcella Cornia, Matteo Stefanini, Lorenzo Baraldi, and Rita Cucchiara. 2020.
\newblock {Meshed-Memory Transformer for Image Captioning}.
\newblock In \emph{CVPR}.

\bibitem[{Dai et~al.(2018)Dai, Fidler, and Lin}]{DBLP:conf/nips/DaiFL18}
Bo~Dai, Sanja Fidler, and Dahua Lin. 2018.
\newblock A neural compositional paradigm for image captioning.
\newblock In \emph{Advances in Neural Information Processing Systems}, pages
  658--668.

\bibitem[{Devlin et~al.(2019)Devlin, Chang, Lee, and
  Toutanova}]{DBLP:conf/naacl/DevlinCLT19}
Jacob Devlin, Ming{-}Wei Chang, Kenton Lee, and Kristina Toutanova. 2019.
\newblock \href {https://doi.org/10.18653/v1/n19-1423} {{BERT:} pre-training of
  deep bidirectional transformers for language understanding}.
\newblock In \emph{Proceedings of the 2019 Conference of the North American
  Chapter of the Association for Computational Linguistics: Human Language
  Technologies, {NAACL-HLT} 2019, Minneapolis, MN, USA, June 2-7, 2019, Volume
  1 (Long and Short Papers)}, pages 4171--4186. Association for Computational
  Linguistics.

\bibitem[{Elliott et~al.(2016)Elliott, Frank, Sima'an, and
  Specia}]{DBLP:conf/acl/ElliottFSS16}
Desmond Elliott, Stella Frank, Khalil Sima'an, and Lucia Specia. 2016.
\newblock \href {https://doi.org/10.18653/v1/w16-3210} {Multi30k: Multilingual
  english-german image descriptions}.
\newblock In \emph{Proceedings of the 5th Workshop on Vision and Language,
  hosted by the 54th Annual Meeting of the Association for Computational
  Linguistics, VL@ACL 2016, August 12, Berlin, Germany}. The Association for
  Computer Linguistics.

\bibitem[{Elliott and Keller(2013)}]{DBLP:conf/emnlp/ElliottK13}
Desmond Elliott and Frank Keller. 2013.
\newblock Image description using visual dependency representations.
\newblock In \emph{Proceedings of the 2013 Conference on Empirical Methods in
  Natural Language Processing}, pages 1292--1302.

\bibitem[{Fang et~al.(2015)Fang, Gupta, Iandola, Srivastava, Deng,
  Doll{\'{a}}r, Gao, He, Mitchell, Platt, Zitnick, and
  Zweig}]{DBLP:conf/cvpr/FangGISDDGHMPZZ15}
Hao Fang, Saurabh Gupta, Forrest~N. Iandola, Rupesh~Kumar Srivastava, Li~Deng,
  Piotr Doll{\'{a}}r, Jianfeng Gao, Xiaodong He, Margaret Mitchell, John~C.
  Platt, C.~Lawrence Zitnick, and Geoffrey Zweig. 2015.
\newblock \href {https://doi.org/10.1109/CVPR.2015.7298754} {From captions to
  visual concepts and back}.
\newblock In \emph{{IEEE} Conference on Computer Vision and Pattern
  Recognition, {CVPR} 2015, Boston, MA, USA, June 7-12, 2015}, pages
  1473--1482. {IEEE} Computer Society.

\bibitem[{Feng et~al.(2019)Feng, Ma, Liu, and Luo}]{DBLP:conf/cvpr/Feng00L19a}
Yang Feng, Lin Ma, Wei Liu, and Jiebo Luo. 2019.
\newblock Unsupervised image captioning.
\newblock In \emph{Proceedings of the IEEE conference on computer vision and
  pattern recognition}, pages 4125--4134.

\bibitem[{Geva et~al.(2019)Geva, Goldberg, and
  Berant}]{DBLP:conf/emnlp/GevaGB19}
Mor Geva, Yoav Goldberg, and Jonathan Berant. 2019.
\newblock \href {https://doi.org/10.18653/v1/D19-1107} {Are we modeling the
  task or the annotator? an investigation of annotator bias in natural language
  understanding datasets}.
\newblock In \emph{Proceedings of the 2019 Conference on Empirical Methods in
  Natural Language Processing and the 9th International Joint Conference on
  Natural Language Processing, {EMNLP-IJCNLP} 2019, Hong Kong, China, November
  3-7, 2019}, pages 1161--1166. Association for Computational Linguistics.

\bibitem[{Gu et~al.(2018)Gu, Joty, Cai, and Wang}]{DBLP:conf/eccv/GuJCW18}
Jiuxiang Gu, Shafiq~R. Joty, Jianfei Cai, and Gang Wang. 2018.
\newblock \href {https://doi.org/10.1007/978-3-030-01246-5\_31} {Unpaired image
  captioning by language pivoting}.
\newblock In \emph{Computer Vision - {ECCV} 2018 - 15th European Conference,
  Munich, Germany, September 8-14, 2018, Proceedings, Part {I}}, volume 11205
  of \emph{Lecture Notes in Computer Science}, pages 519--535. Springer.

\bibitem[{Gu et~al.(2019)Gu, Joty, Cai, Zhao, Yang, and
  Wang}]{DBLP:conf/iccv/GuJCZYW19}
Jiuxiang Gu, Shafiq~R. Joty, Jianfei Cai, Handong Zhao, Xu~Yang, and Gang Wang.
  2019.
\newblock \href {https://doi.org/10.1109/ICCV.2019.01042} {Unpaired image
  captioning via scene graph alignments}.
\newblock In \emph{2019 {IEEE/CVF} International Conference on Computer Vision,
  {ICCV} 2019, Seoul, Korea (South), October 27 - November 2, 2019}, pages
  10322--10331. {IEEE}.

\bibitem[{Guo et~al.(2019)Guo, Liu, Yao, Li, and Lu}]{guo2019mscap}
Longteng Guo, Jing Liu, Peng Yao, Jiangwei Li, and Hanqing Lu. 2019.
\newblock Mscap: Multi-style image captioning with unpaired stylized text.
\newblock In \emph{Proceedings of the IEEE Conference on Computer Vision and
  Pattern Recognition}, pages 4204--4213.

\bibitem[{He et~al.(2016)He, Zhang, Ren, and Sun}]{resnet}
Kaiming He, Xiangyu Zhang, Shaoqing Ren, and Jian Sun. 2016.
\newblock Deep residual learning for image recognition.
\newblock In \emph{CVPR}.

\bibitem[{Hinton et~al.(2015)Hinton, Vinyals, and Dean}]{hinton2015distilling}
Geoffrey Hinton, Oriol Vinyals, and Jeff Dean. 2015.
\newblock Distilling the knowledge in a neural network.
\newblock In \emph{Proceedings of NIPS workshop}.

\bibitem[{Hossain et~al.(2018)Hossain, Sohel, Shiratuddin, and
  Laga}]{DBLP:journals/corr/abs-1810-04020}
Md.~Zakir Hossain, Ferdous Sohel, Mohd~Fairuz Shiratuddin, and Hamid Laga.
  2018.
\newblock \href {http://arxiv.org/abs/1810.04020} {A comprehensive survey of
  deep learning for image captioning}.
\newblock \emph{CoRR}, abs/1810.04020.

\bibitem[{Huang et~al.(2019)Huang, Wang, Chen, and Wei}]{huang19attention}
Lun Huang, Wenmin Wang, Jie Chen, and Xiao-Yong Wei. 2019.
\newblock Attention on attention for image captioning.
\newblock In \emph{CVPR}.

\bibitem[{Juan et~al.(2019)Juan, Lu, Li, Peng, Timofeev, Chen, Gao, Duerig,
  Tomkins, and Ravi}]{DBLP:journals/corr/abs-1902-10814}
Da{-}Cheng Juan, Chun{-}Ta Lu, Zhen Li, Futang Peng, Aleksei Timofeev,
  Yi{-}Ting Chen, Yaxi Gao, Tom Duerig, Andrew Tomkins, and Sujith Ravi. 2019.
\newblock \href {http://arxiv.org/abs/1902.10814} {Graph-rise:
  Graph-regularized image semantic embedding}.
\newblock \emph{CoRR}, abs/1902.10814.

\bibitem[{Juan et~al.(2020)Juan, Lu, Li, Peng, Timofeev, Chen, Gao, Duerig,
  Tomkins, and Ravi}]{DBLP:conf/wsdm/JuanLLPTCGDTR20}
Da-Cheng Juan, Chun-Ta Lu, Zhen Li, Futang Peng, Aleksei Timofeev, Yi-Ting
  Chen, Yaxi Gao, Tom Duerig, Andrew Tomkins, and Sujith Ravi. 2020.
\newblock Ultra fine-grained image semantic embedding.
\newblock In \emph{Proceedings of the 13th International Conference on Web
  Search and Data Mining}, pages 277--285.

\bibitem[{Kilickaya et~al.(2017)Kilickaya, Erdem, Ikizler{-}Cinbis, and
  Erdem}]{DBLP:conf/eacl/KilickayaEIE17}
Mert Kilickaya, Aykut Erdem, Nazli Ikizler{-}Cinbis, and Erkut Erdem. 2017.
\newblock \href {https://doi.org/10.18653/v1/e17-1019} {Re-evaluating automatic
  metrics for image captioning}.
\newblock In \emph{Proceedings of the 15th Conference of the European Chapter
  of the Association for Computational Linguistics, {EACL} 2017, Valencia,
  Spain, April 3-7, 2017, Volume 1: Long Papers}, pages 199--209. Association
  for Computational Linguistics.

\bibitem[{Kim et~al.(2019)Kim, Choi, Hwang, Song, Lee, Woo, and
  Modulabs}]{kimvizwiz}
Suwon Kim, HongYong Choi, JoongWon Hwang, JangYoung Song, SangRok Lee, TaeKang
  Woo, and AI~Modulabs. 2019.
\newblock Vizwiz image captioning based on aoanet with scene graph.
\newblock \emph{ivc.ischool.utexas.edu}.

\bibitem[{Krishna et~al.(2017)Krishna, Zhu, Groth, Johnson, Hata, Kravitz,
  Chen, Kalantidis, Li, Shamma, Bernstein, and Fei-Fei}]{krishnavisualgenome}
Ranjay Krishna, Yuke Zhu, Oliver Groth, Justin Johnson, Kenji Hata, Joshua
  Kravitz, Stephanie Chen, Yannis Kalantidis, Li-Jia Li, David~A. Shamma,
  Michael Bernstein, and Li~Fei-Fei. 2017.
\newblock {Visual Genome}: Connecting language and vision using crowdsourced
  dense image annotations.
\newblock \emph{IJCV}, 123(1):32--73.

\bibitem[{Kulkarni et~al.(2013)Kulkarni, Premraj, Ordonez, Dhar, Li, Choi,
  Berg, and Berg}]{DBLP:conf/cvpr/KulkarniPDLCBB11}
Girish Kulkarni, Visruth Premraj, Vicente Ordonez, Sagnik Dhar, Siming Li,
  Yejin Choi, Alexander~C Berg, and Tamara~L Berg. 2013.
\newblock Babytalk: Understanding and generating simple image descriptions.
\newblock volume~35, pages 2891--2903. IEEE.

\bibitem[{Kuznetsova et~al.(2020)Kuznetsova, Rom, Alldrin, Uijlings, Krasin,
  Pont-Tuset, Kamali, Popov, Malloci, Kolesnikov
  et~al.}]{DBLP:journals/corr/abs-1811-00982}
Alina Kuznetsova, Mohamad Hassan~Mohamad Rom, Neil Alldrin, Jasper Uijlings,
  Ivan Krasin, Jordi Pont-Tuset, Shahab Kamali, Stefan Popov, Matteo Malloci,
  Alexander Kolesnikov, et~al. 2020.
\newblock The open images dataset v4: Unified image classification, object
  detection, and visual relationship detection at scale.

\bibitem[{Kuznetsova et~al.(2014)Kuznetsova, Ordonez, Berg, and
  Choi}]{kuznetsova2014treetalk}
Polina Kuznetsova, Vicente Ordonez, Tamara~L Berg, and Yejin Choi. 2014.
\newblock Treetalk: Composition and compression of trees for image
  descriptions.
\newblock \emph{Transactions of the Association for Computational Linguistics},
  2:351--362.

\bibitem[{Lan et~al.(2017)Lan, Li, and Dong}]{lan2017fluency}
Weiyu Lan, Xirong Li, and Jianfeng Dong. 2017.
\newblock Fluency-guided cross-lingual image captioning.
\newblock In \emph{Proceedings of the 25th ACM international conference on
  Multimedia}, pages 1549--1557.

\bibitem[{Latcinnik and Berant(2020)}]{DBLP:journals/corr/abs-2004-05569}
Veronica Latcinnik and Jonathan Berant. 2020.
\newblock \href {http://arxiv.org/abs/2004.05569} {Explaining question
  answering models through text generation}.
\newblock \emph{CoRR}, abs/2004.05569.

\bibitem[{Li et~al.(2020)Li, Duan, Fang, Jiang, and Zhou}]{li2020unicoder}
Gen Li, Nan Duan, Yuejian Fang, Daxin Jiang, and Ming Zhou. 2020.
\newblock {Unicoder-VL}: A universal encoder for vision and language by
  cross-modal pre-training.
\newblock In \emph{AAAI}.

\bibitem[{Li et~al.(2019{\natexlab{a}})Li, Zhu, Liu, and
  Yang}]{DBLP:conf/iccv/LiZLY19}
Guang Li, Linchao Zhu, Ping Liu, and Yi~Yang. 2019{\natexlab{a}}.
\newblock \href {https://doi.org/10.1109/ICCV.2019.00902} {Entangled
  transformer for image captioning}.
\newblock In \emph{2019 {IEEE/CVF} International Conference on Computer Vision,
  {ICCV} 2019, Seoul, Korea (South), October 27 - November 2, 2019}, pages
  8927--8936. {IEEE}.

\bibitem[{Li et~al.(2019{\natexlab{b}})Li, Yao, Guo, and Zhang}]{li2019boosted}
Jiangyun Li, Peng Yao, Longteng Guo, and Weicun Zhang. 2019{\natexlab{b}}.
\newblock Boosted transformer for image captioning.
\newblock \emph{Applied Sciences (2076-3417)}, 9(16).

\bibitem[{Li et~al.(2022)Li, Li, Xiong, and
  Hoi}]{DBLP:journals/corr/abs-2201-12086}
Junnan Li, Dongxu Li, Caiming Xiong, and Steven C.~H. Hoi. 2022.
\newblock \href {http://arxiv.org/abs/2201.12086} {{BLIP:} bootstrapping
  language-image pre-training for unified vision-language understanding and
  generation}.
\newblock \emph{CoRR}, abs/2201.12086.

\bibitem[{Li and Chen(2018)}]{DBLP:conf/ijcai/LiC18}
Nannan Li and Zhenzhong Chen. 2018.
\newblock \href {https://doi.org/10.24963/ijcai.2018/110} {Image cationing with
  visual-semantic {LSTM}}.
\newblock In \emph{Proceedings of the Twenty-Seventh International Joint
  Conference on Artificial Intelligence, {IJCAI} 2018, July 13-19, 2018,
  Stockholm, Sweden}, pages 793--799. ijcai.org.

\bibitem[{Li et~al.(2011)Li, Kulkarni, Berg, Berg, and Choi}]{li2011composing}
Siming Li, Girish Kulkarni, Tamara Berg, Alexander Berg, and Yejin Choi. 2011.
\newblock Composing simple image descriptions using web-scale n-grams.
\newblock In \emph{Proceedings of the Fifteenth Conference on Computational
  Natural Language Learning}, pages 220--228.

\bibitem[{Lin et~al.(2014)Lin, Maire, Belongie, Hays, Perona, Ramanan,
  Doll{\'a}r, and Zitnick}]{DBLP:conf/eccv/LinMBHPRDZ14}
Tsung-Yi Lin, Michael Maire, Serge Belongie, James Hays, Pietro Perona, Deva
  Ramanan, Piotr Doll{\'a}r, and C~Lawrence Zitnick. 2014.
\newblock Microsoft coco: Common objects in context.
\newblock In \emph{European conference on computer vision}, pages 740--755.
  Springer.

\bibitem[{Lipton(2018)}]{DBLP:journals/cacm/Lipton18}
Zachary~C. Lipton. 2018.
\newblock \href {https://doi.org/10.1145/3233231} {The mythos of model
  interpretability}.
\newblock \emph{Commun. {ACM}}, 61(10):36--43.

\bibitem[{Liu et~al.(2018)Liu, Ren, Liu, Wang, and Sun}]{liu2018simnet}
Fenglin Liu, Xuancheng Ren, Yuanxin Liu, Houfeng Wang, and Xu~Sun. 2018.
\newblock simnet: Stepwise image-topic merging network for generating detailed
  and comprehensive image captions.
\newblock In \emph{Proceedings of the 2018 Conference on Empirical Methods in
  Natural Language Processing}, pages 137--149.

\bibitem[{Lu et~al.(2019{\natexlab{a}})Lu, Batra, Parikh, and
  Lee}]{lu19vilbert}
Jiasen Lu, Dhruv Batra, Devi Parikh, and Stefan Lee. 2019{\natexlab{a}}.
\newblock {ViLBERT}: Pretraining task-agnostic visiolinguistic representations
  for vision-and-language tasks.
\newblock In \emph{NeurIPS}.

\bibitem[{Lu et~al.(2019{\natexlab{b}})Lu, Batra, Parikh, and
  Lee}]{lu2019vilbert}
Jiasen Lu, Dhruv Batra, Devi Parikh, and Stefan Lee. 2019{\natexlab{b}}.
\newblock {ViLBERT}: Pretraining task-agnostic visiolinguistic representations
  for vision-and-language tasks.
\newblock In \emph{NeurIPS}.

\bibitem[{Lu et~al.(2019{\natexlab{c}})Lu, Batra, Parikh, and
  Lee}]{DBLP:conf/nips/LuBPL19}
Jiasen Lu, Dhruv Batra, Devi Parikh, and Stefan Lee. 2019{\natexlab{c}}.
\newblock \href
  {https://proceedings.neurips.cc/paper/2019/hash/c74d97b01eae257e44aa9d5bade97baf-Abstract.html}
  {Vilbert: Pretraining task-agnostic visiolinguistic representations for
  vision-and-language tasks}.
\newblock In \emph{Advances in Neural Information Processing Systems 32: Annual
  Conference on Neural Information Processing Systems 2019, NeurIPS 2019,
  December 8-14, 2019, Vancouver, BC, Canada}, pages 13--23.

\bibitem[{Luo and Shakhnarovich(2020)}]{DBLP:journals/corr/abs-2005-14386}
Ruotian Luo and Greg Shakhnarovich. 2020.
\newblock \href {http://arxiv.org/abs/2005.14386} {Controlling length in image
  captioning}.
\newblock \emph{CoRR}, abs/2005.14386.

\bibitem[{Mathews et~al.(2018)Mathews, Xie, and He}]{mathews2018semstyle}
Alexander Mathews, Lexing Xie, and Xuming He. 2018.
\newblock Semstyle: Learning to generate stylised image captions using
  unaligned text.
\newblock In \emph{Proceedings of the IEEE Conference on Computer Vision and
  Pattern Recognition}, pages 8591--8600.

\bibitem[{Plummer et~al.(2015)Plummer, Wang, Cervantes, Caicedo, Hockenmaier,
  and Lazebnik}]{plummer2015flickr30k}
Bryan~A Plummer, Liwei Wang, Chris~M Cervantes, Juan~C Caicedo, Julia
  Hockenmaier, and Svetlana Lazebnik. 2015.
\newblock Flickr30k entities: Collecting region-to-phrase correspondences for
  richer image-to-sentence models.
\newblock In \emph{Proceedings of the IEEE international conference on computer
  vision}, pages 2641--2649.

\bibitem[{Ren et~al.(2015)Ren, He, Girshick, and Sun}]{ren2015faster}
Shaoqing Ren, Kaiming He, Ross~B. Girshick, and Jian Sun. 2015.
\newblock \href
  {http://papers.nips.cc/paper/5638-faster-r-cnn-towards-real-time-object-detection-with-region-proposal-networks}
  {Faster {R-CNN:} towards real-time object detection with region proposal
  networks}.
\newblock In \emph{Advances in Neural Information Processing Systems 28: Annual
  Conference on Neural Information Processing Systems 2015, December 7-12,
  2015, Montreal, Quebec, Canada}, pages 91--99.

\bibitem[{Russakovsky et~al.(2015)Russakovsky, Deng, Su, Krause, Satheesh, Ma,
  Huang, Karpathy, Khosla, Bernstein, Berg, and Fei-Fei}]{imagenet15}
Olga Russakovsky, Jia Deng, Hao Su, Jonathan Krause, Sanjeev Satheesh, Sean Ma,
  Zhiheng Huang, Andrej Karpathy, Aditya Khosla, Michael Bernstein,
  Alexander~C. Berg, and Li~Fei-Fei. 2015.
\newblock {ImageNet} large scale visual recognition challenge.
\newblock \emph{IJCV}, 115(3):211--252.

\bibitem[{Seo et~al.(2020)Seo, Sharma, Levinboim, Han, and
  Soricut}]{DBLP:conf/aaai/SeoSLHS20}
Paul~Hongsuck Seo, Piyush Sharma, Tomer Levinboim, Bohyung Han, and Radu
  Soricut. 2020.
\newblock \href {https://aaai.org/ojs/index.php/AAAI/article/view/5655}
  {Reinforcing an image caption generator using off-line human feedback}.
\newblock In \emph{The Thirty-Fourth {AAAI} Conference on Artificial
  Intelligence, {AAAI} 2020, The Thirty-Second Innovative Applications of
  Artificial Intelligence Conference, {IAAI} 2020, The Tenth {AAAI} Symposium
  on Educational Advances in Artificial Intelligence, {EAAI} 2020, New York,
  NY, USA, February 7-12, 2020}, pages 2693--2700. {AAAI} Press.

\bibitem[{Sharma et~al.(2020)Sharma, Agrahari, Singh, Firoj, and
  Mishra}]{sharma2020image}
Himanshu Sharma, Manmohan Agrahari, Sujeet~Kumar Singh, Mohd Firoj, and
  Ravi~Kumar Mishra. 2020.
\newblock Image captioning: A comprehensive survey.
\newblock In \emph{2020 International Conference on Power Electronics \& IoT
  Applications in Renewable Energy and its Control (PARC)}, pages 325--328.
  IEEE.

\bibitem[{Sharma et~al.(2018)Sharma, Ding, Goodman, and
  Soricut}]{DBLP:conf/acl/SoricutDSG18}
Piyush Sharma, Nan Ding, Sebastian Goodman, and Radu Soricut. 2018.
\newblock Conceptual captions: A cleaned, hypernymed, image alt-text dataset
  for automatic image captioning.
\newblock In \emph{Proceedings of the 56th Annual Meeting of the Association
  for Computational Linguistics (Volume 1: Long Papers)}, pages 2556--2565.

\bibitem[{Su et~al.(2020)Su, Zhu, Cao, Li, Lu, Wei, and Dai}]{su20vlbert}
Weijie Su, Xizhou Zhu, Yue Cao, Bin Li, Lewei Lu, Furu Wei, and Jifeng Dai.
  2020.
\newblock {VL-BERT}: Pre-training of generic visual-linguistic representations.
\newblock In \emph{ICLR}.

\bibitem[{Tan and Bansal(2019)}]{tan19lxmert}
Hao Tan and Mohit Bansal. 2019.
\newblock {LXMERT}: Learning cross-modality encoder representations from
  transformers.
\newblock In \emph{EMNLP-IJCNLP}.

\bibitem[{Tan and Chan(2016)}]{DBLP:conf/accv/TanC16}
Ying~Hua Tan and Chee~Seng Chan. 2016.
\newblock \href {https://doi.org/10.1007/978-3-319-54193-8\_7} {phi-lstm: {A}
  phrase-based hierarchical {LSTM} model for image captioning}.
\newblock In \emph{Computer Vision - {ACCV} 2016 - 13th Asian Conference on
  Computer Vision, Taipei, Taiwan, November 20-24, 2016, Revised Selected
  Papers, Part {V}}, volume 10115 of \emph{Lecture Notes in Computer Science},
  pages 101--117. Springer.

\bibitem[{Thapliyal and Soricut(2020)}]{DBLP:conf/acl/ThapliyalS20}
Ashish~V. Thapliyal and Radu Soricut. 2020.
\newblock \href {https://www.aclweb.org/anthology/2020.acl-main.16/}
  {Cross-modal language generation using pivot stabilization for web-scale
  language coverage}.
\newblock In \emph{Proceedings of the 58th Annual Meeting of the Association
  for Computational Linguistics, {ACL} 2020, Online, July 5-10, 2020}, pages
  160--170. Association for Computational Linguistics.

\bibitem[{Thorne et~al.(2019)Thorne, Vlachos, Christodoulopoulos, and
  Mittal}]{DBLP:conf/naacl/ThorneVCM19}
James Thorne, Andreas Vlachos, Christos Christodoulopoulos, and Arpit Mittal.
  2019.
\newblock \href {https://doi.org/10.18653/v1/n19-1101} {Generating token-level
  explanations for natural language inference}.
\newblock In \emph{Proceedings of the 2019 Conference of the North American
  Chapter of the Association for Computational Linguistics: Human Language
  Technologies, {NAACL-HLT} 2019, Minneapolis, MN, USA, June 2-7, 2019, Volume
  1 (Long and Short Papers)}, pages 963--969. Association for Computational
  Linguistics.

\bibitem[{Tsuchiya(2018)}]{DBLP:conf/lrec/Tsuchiya18}
Masatoshi Tsuchiya. 2018.
\newblock \href
  {http://www.lrec-conf.org/proceedings/lrec2018/summaries/786.html}
  {Performance impact caused by hidden bias of training data for recognizing
  textual entailment}.
\newblock In \emph{Proceedings of the Eleventh International Conference on
  Language Resources and Evaluation, {LREC} 2018, Miyazaki, Japan, May 7-12,
  2018}. European Language Resources Association {(ELRA)}.

\bibitem[{Tsutsui and Crandall(2017)}]{tsutsui2017using}
Satoshi Tsutsui and David Crandall. 2017.
\newblock Using artificial tokens to control languages for multilingual image
  caption generation.
\newblock \emph{arXiv preprint arXiv:1706.06275}.

\bibitem[{Vaswani et~al.(2017)Vaswani, Shazeer, Parmar, Uszkoreit, Jones,
  Gomez, Kaiser, and Polosukhin}]{DBLP:conf/nips/VaswaniSPUJGKP17}
Ashish Vaswani, Noam Shazeer, Niki Parmar, Jakob Uszkoreit, Llion Jones,
  Aidan~N Gomez, {\L}ukasz Kaiser, and Illia Polosukhin. 2017.
\newblock Attention is all you need.
\newblock In \emph{Advances in neural information processing systems}, pages
  5998--6008.

\bibitem[{Wang et~al.(2019{\natexlab{a}})Wang, Beck, and Cohn}]{wang2019role}
Dalin Wang, Daniel Beck, and Trevor Cohn. 2019{\natexlab{a}}.
\newblock On the role of scene graphs in image captioning.
\newblock In \emph{Proceedings of the Beyond Vision and LANguage: inTEgrating
  Real-world kNowledge (LANTERN)}, pages 29--34.

\bibitem[{Wang et~al.(2019{\natexlab{b}})Wang, Zhao, Yatskar, Chang, and
  Ordonez}]{wang2019balanced}
Tianlu Wang, Jieyu Zhao, Mark Yatskar, Kai-Wei Chang, and Vicente Ordonez.
  2019{\natexlab{b}}.
\newblock Balanced datasets are not enough: Estimating and mitigating gender
  bias in deep image representations.
\newblock In \emph{Proceedings of the IEEE International Conference on Computer
  Vision}, pages 5310--5319.

\bibitem[{Wang et~al.(2017)Wang, Lin, Shen, Cohen, and
  Cottrell}]{wang2017skeleton}
Yufei Wang, Zhe Lin, Xiaohui Shen, Scott Cohen, and Garrison~W Cottrell. 2017.
\newblock Skeleton key: Image captioning by skeleton-attribute decomposition.
\newblock In \emph{Proceedings of the IEEE conference on computer vision and
  pattern recognition}, pages 7272--7281.

\bibitem[{Wiegreffe and Pinter(2019)}]{DBLP:conf/emnlp/WiegreffeP19}
Sarah Wiegreffe and Yuval Pinter. 2019.
\newblock \href {https://doi.org/10.18653/v1/D19-1002} {Attention is not not
  explanation}.
\newblock In \emph{Proceedings of the 2019 Conference on Empirical Methods in
  Natural Language Processing and the 9th International Joint Conference on
  Natural Language Processing, {EMNLP-IJCNLP} 2019, Hong Kong, China, November
  3-7, 2019}, pages 11--20. Association for Computational Linguistics.

\bibitem[{Xu et~al.(2015)Xu, Ba, Kiros, Cho, Courville, Salakhudinov, Zemel,
  and Bengio}]{xu2015show}
Kelvin Xu, Jimmy Ba, Ryan Kiros, Kyunghyun Cho, Aaron Courville, Ruslan
  Salakhudinov, Rich Zemel, and Yoshua Bengio. 2015.
\newblock Show, attend and tell: Neural image caption generation with visual
  attention.
\newblock In \emph{International conference on machine learning}, pages
  2048--2057.

\bibitem[{Yang et~al.(2019)Yang, Tang, Zhang, and
  Cai}]{DBLP:conf/cvpr/YangTZC19}
Xu~Yang, Kaihua Tang, Hanwang Zhang, and Jianfei Cai. 2019.
\newblock Auto-encoding scene graphs for image captioning.
\newblock In \emph{Proceedings of the IEEE Conference on Computer Vision and
  Pattern Recognition}, pages 10685--10694.

\bibitem[{Yao et~al.(2018)Yao, Pan, Li, and Mei}]{DBLP:conf/eccv/YaoPLM18}
Ting Yao, Yingwei Pan, Yehao Li, and Tao Mei. 2018.
\newblock Exploring visual relationship for image captioning.
\newblock In \emph{Proceedings of the European conference on computer vision
  (ECCV)}, pages 684--699.

\bibitem[{Yao et~al.(2017)Yao, Pan, Li, Qiu, and Mei}]{yao2017boosting}
Ting Yao, Yingwei Pan, Yehao Li, Zhaofan Qiu, and Tao Mei. 2017.
\newblock Boosting image captioning with attributes.
\newblock In \emph{Proceedings of the IEEE International Conference on Computer
  Vision}, pages 4894--4902.

\bibitem[{Yoshikawa et~al.(2017)Yoshikawa, Shigeto, and
  Takeuchi}]{DBLP:conf/acl/YoshikawaST17}
Yuya Yoshikawa, Yutaro Shigeto, and Akikazu Takeuchi. 2017.
\newblock \href {https://doi.org/10.18653/v1/P17-2066} {{STAIR} captions:
  Constructing a large-scale japanese image caption dataset}.
\newblock In \emph{Proceedings of the 55th Annual Meeting of the Association
  for Computational Linguistics, {ACL} 2017, Vancouver, Canada, July 30 -
  August 4, Volume 2: Short Papers}, pages 417--421. Association for
  Computational Linguistics.

\bibitem[{You et~al.(2016)You, Jin, Wang, Fang, and Luo}]{you2016image}
Quanzeng You, Hailin Jin, Zhaowen Wang, Chen Fang, and Jiebo Luo. 2016.
\newblock Image captioning with semantic attention.
\newblock In \emph{Proceedings of the IEEE conference on computer vision and
  pattern recognition}, pages 4651--4659.

\bibitem[{Yu et~al.(2019)Yu, Li, Yu, and Huang}]{yu2019multimodal}
Jun Yu, Jing Li, Zhou Yu, and Qingming Huang. 2019.
\newblock Multimodal transformer with multi-view visual representation for
  image captioning.
\newblock \emph{arXiv}, 1905.07841.

\bibitem[{Zheng et~al.(2019)Zheng, Li, and Wang}]{DBLP:conf/cvpr/ZhengLW19}
Yue Zheng, Yali Li, and Shengjin Wang. 2019.
\newblock \href {https://doi.org/10.1109/CVPR.2019.00859} {Intention oriented
  image captions with guiding objects}.
\newblock In \emph{{IEEE} Conference on Computer Vision and Pattern
  Recognition, {CVPR} 2019, Long Beach, CA, USA, June 16-20, 2019}, pages
  8395--8404. Computer Vision Foundation / {IEEE}.

\end{thebibliography}
\bibliographystyle{acl_natbib}

\clearpage
\pagebreak
\appendix

\section{Comparison of SkeEnc and SkeAE on multilingual captions}
\label{sec:app_multilingual}

We have discussed the human evaluation scores of the SkeAE model by using \textit{nouns and verbs} as skeletons in Table \ref{tab:he_multilingual} in the main paper. In addition to this, we also conducted human evaluation to compare the SkeEnc model with the \textit{nouns and verbs} skeletons in comparison to the baseline. We present this in Table \ref{tab:he_multilingual_SkeEnc}. While there are improvements in 3 languages, the performance is also hurt in two languages. However, as we see, by comparing the performances in Table \ref{tab:he_multilingual} and Table \ref{tab:he_multilingual_SkeEnc}, we observe that SkeAE has a clear advantage when leveraging the English caption to improve multilingual captions. This clearly indicates that channelling the prediction of the skeleton words in conjuction with the caption itself is enabling the model decoder to attend to the previously predicted skeleton words in the same decoder. 

\begin{table}[h]
\centering
\resizebox{0.28\textwidth}{!}{%
\begin{tabular}{l|l|l|l}
\hline \hline
\textbf{Language} & \textbf{Wins} & \textbf{Losses} & \textbf{Gains} \\ \hline
\textbf{French} & 31.93 & 31.43 & \textbf{+0.50} \\
\textbf{Italian} & 33.13 & 28.32 & \textbf{+4.81} \\
\textbf{German} & 29.43 & 29.72 & \textbf{-0.30} \\
\textbf{Spanish} & 30.53 & 34.43 & -3.90 \\
\textbf{Hindi} & 29.93 & 26.03 & \textbf{+3.90} \\ \hline \hline
\end{tabular}
}
%\vspace{-6pt}
\caption{\small{Human evaluation results on SkeEnc model for skeleton (form: nouns \& verbs, prediction approach: generation) conditioned caption generation for multiple languages.}}
%\vspace{-10pt}
\label{tab:he_multilingual_SkeEnc}
\end{table}

\section{Comparison of Classification and Generation based Skeleton Prediction}
%SkeEnc
From a preliminary manual analysis, we observed that the classification based approach to skeleton prediction faces the problem of predicting words that are related but are not likely to co-occur within the same sentence in the caption. This is described in detail in points 1a and 1b of \S  \ref{sec:method}. To validate this observation, we conducted human evaluation of the captions generated from classification and generation based approaches relative to one another. This setup is different from the rest of the experiments in human evaluation in the paper which compare any given model relative to the baseline model. In contrast, this study is to compare the generation and classification approaches with one another. These results are presented in Table \ref{tab:he_gen_vs_clf}.

The top-8 highest scoring content words are chosen to reduce input noise for the caption generator while improving the recall of concepts. We experimented with different values for this and selected 8 to be an optimal balance between the content in the skeleton words and the noise.

\begin{table}[h]
\centering
\resizebox{0.28\textwidth}{!}{%
\begin{tabular}{l|l|l|l}
\hline \hline
\textbf{Approach} & \textbf{Wins} & \textbf{Losses} & \textbf{Gains} \\ \hline
\textbf{Generation} & 39.14 & 30.23 & \textbf{+8.91} \\
\hline \hline
\end{tabular}
}
%\vspace{-6pt}
\caption{\small{Human evaluation results of comparison between the generation and classification based approaches}}
%\vspace{-10pt}
\label{tab:he_gen_vs_clf}
\end{table}

We observe that the generation based approach has significant gains of +8.91 over the classification based approach. Most of the prior literature uses the classification based approach to predict content or bag of concepts to assist caption generation. Our hypothesis is that this classification based model helps in end-to-end approaches where the loss from caption generation backpropagates to the classifier model as well. As opposed to this, our model decouples the prediction of the skeleton or concept words that are further used for caption generation. Hence we believe that suppressing the words that do not co-occur is important in the skeleton prediction task and the generation based approach is addressing this problem.

\section{Absolute Ratings}

Here are some of the observations from these results:
\begin{itemize}
    \item \textit{Better results of Dual Staged Approach: }As we can see in the last two rows (rows 8 and 9), our proposed SkeEnc and SkeAE show absolute improvements in both the categories. This further demonstrates that the proposed dual staged approach is generating better denoised captions when trained on noisy uncurated alt-text--based captions.
    \item \textit{Sub-selecting content words is better: } Now that we saw the improvements with the dual staged approach, we now investigate whether sub-selecting content words is important. For this, we present comparison between rows 7 and 8. Both these models are dual staged with SkeEnc i.e encoding the predicted skeleton in the second stage. The only difference is that row 8 sub-selects all nouns and verbs to predict the skeletons whereas row 8 includes all the words from the captions to predict the skeletons. Row 8 shows better performance compared to row 7. This means that sub-selecting content words contribute to the caption generation in the second stage.
\end{itemize}

Please note that we focus on alt-text based captions, so we experiment on Conceptual Captions instead of cleaner alternatives such as MSCOCO and Multi30k. The latter do not include as noisy captions as they are hand-annotated (refer Section \ref{sec:datasets})

% 
% These ratings are collected to double check the results (we do not make further use of them in the paper).

\section{Img2Ske: Classification based prediction}
\label{sec:classification}
% The skeleton words are treated as independent labels in the classification approach. 
Skeleton prediction is posed as a multilabel classification problem where the prediction of a skeleton word $s_i$ is not conditionally dependent on the prediction of another skeleton word $s_j$. Our goal is to evaluate the effectiveness of simple generation and classification models to predict skeletons, and naturally generation based approach reduces redundancies due to conditional dependence of label/skeleton prediction. 
The encoder part remains the same as the baseline followed by
%This approach has the advantage of gaining more freedom by predicting skeleton words conditionally independent of one another.
optimization with sigmoid cross entropy between the skeleton words $\mathbb{S}$ and image encoding $\boldsymbol{z_{\mathbb{I}}}$, which is the representation of the image from the encoder.
% \[
% \small
% - [ \mathbb{S} \cdot \log (\sigma(z_{\mathbb{I}}))+ (1-\mathbb{S}) \cdot \log (1-\sigma(z_{\mathbb{I}})) ]
% \]

\begin{equation}
\text {Accuracy, } A=\frac{1}{N} \sum_{i=1}^{N} \frac{\left|\mathbb{S}_{i} \cap \mathbb{\hat{S}}_{i}\right|}{\left|\mathbb{S}_{i} \cup \mathbb{\hat{S}}_{i}\right|}
\end{equation}

%Model selection is performed on the basis of accuracy computed between the many-hot representation of the ground truth skeleton words and the predicted skeleton words. 
\noindent The skeleton for the second stage is chosen as the ordered list of top-8 (experimentally selected) high scoring words after the softmax layer. 
%We experimented with different values of k and present the results for $k=8$ in this paper. 
% This presents the skeleton as an ordered set of words. 
However, conditional independence of skeleton words with one another ignores the co-occurrences of words capable of composing a sentence or a final caption. 
For instance, classification predictions are composed of words and their synonyms that are highly correlated like \{\textit{person}, \textit{man}, \textit{singer}\}. 
These words definitely are relevant to an image but do not all necessarily co-occur in a sentence.

Table \ref{tab:clfvsgen} presents the precision, recall and f-scores of the generation and classification based approaches for skeleton prediction. 
These metrics, however are misleading because they do not account for synonyms or semantic similarity. 
For example, `food',  `meal', `lunch' and `dinner' 
are all distinct labels while computing these metrics, and predicting one instead of the other get heavily penalized even though the effect on downstream caption quality would be minimal.
This issue gets amplified by the fact that 
with CC that has a rich vocabulary with words such as electricity `pylon' and `tower' referring to the same concept.

\section{Performance drop for Spanish}

While we have seen improvements in the performance on multiple languages in human evaluation (Table \ref{tab:he_eng}), we observed a drop in the preference for Spanish captions when we use skeletons. 
Given the similarity in word order between Spanish and English in comparison to Hindi, the lower performance of Spanish is an interesting result indeed. Our speculation for this is probably due to the dialect differences. The translation model that we used for Spanish is a mix of `Spain Spanish' and `Latin American Spanish', with Latin American Spanish dominating. The evaluation was done by raters from Spain. The dialects are sufficiently different that it would impact the absolute scores.

\section{Intuition for skeleton words: }

The alt-text captions are silver standard and harbor a lot of diversity. Hence filtering frequently occurring words based on a frequency cutoff as the skeletons helps balance between conditioning on the frequent words (not noise) and diverse concepts. Qualitatively, consider an image of a house with the caption \textit{`apartment for rent'} and \textit{`apartment for sale'}. With the frequency based skeleton selection, the noun word \textit{`apartment'} is selected as skeleton ignoring the rest. In this way, we are denoising alt-text captions to generate captions with visible concepts.

\section{Hyperparameters:}

This section lists the hyperparameters used for training our models. We used BERT embeddings \cite{DBLP:conf/naacl/DevlinCLT19} to initialize the words in skeletons in the SkeEnc and SkeAE models. 
\begin{itemize}
    \item \textit{Learning rate:} We experimented with 3.2\textit{$e^{-5}$}, 0.5, 1, 1.5 and 2 as the learning rate. The experiments presented in the paper have the learning rate of 1. The learning rate is decayed at 0.95 decay rate with staircase strategy.
    \item \textit{Number of layers: } All our models have 6 layers for encoder and decoder. We also conducted an additional experiment to check if the model complexity of the end-to-end baseline can improve the performance in comparison to our dual staged approach. To evaluate this, we doubled the number of layers where the number of transformer encoder and decoder layers are 12 each as presented in the paper as Impr Img2Cap (large) in Table \ref{tab:results_quant} in Section \ref{sec:automatic_metrics}.
    \item \textit{Subtoken Vocabulary:} We experimented with 4000 and 8300 sub-token vocabularies. The experiments in the paper all have 8,300 as subtoken vocabulary size.
    \item \textit{Batch size: } All our experiments include batchsize of 128 only.
    \item \textit{Number of steps: } We train for a maximum of 1 million update steps.
    \item \textit{Maximum Caption Length: } In the baseline and the SkeEnc models, our decoder generates a maximum words of length 36. In the SkeAE and SkeDec model, the skeleton words are prepended to the caption. So we allow the decoder to generate 72 words in these two models.
    \item \textit{Warm up and decay steps: } The model is warmed up for 20 epochs and decayed for 25 epochs.
    \item \textit{Embedding size: } We use embedding dimension of 512.
    \item \textit{Beam size: } We perform beam search in the decoder with a beam size of 5.
\end{itemize}

Here are some of the configuration and modeling choices for training the models:
\begin{itemize}
\item \textit{Attention type: } Our experiments include attention types of cross-attention and text-as-side as described along with point \textit{2a} in Section \ref{sec:method}.
\item \textit{FRCNN Tokens: } We use 1601 tokens from the trained FRCNN.
\end{itemize}

\vfill\null
%\columnbreak

\section{Broader Impact}
\label{sec:broader_impact}
We believe that this work has extensive impact in scaling captioning models to large and noisy datasets thereby exploiting web data and reduce manual annotation efforts.
We do not foresee any immediate concerns ethically directly from our work. 
% However, while applying this to datasets that users are crawling from the web by themselves, we cautiously advise on stripping the data from any offensive and hateful or toxic content. 
% This is not to be overlooked as web-crawled data easily inherits such intents that need to be carefully and definitely filtered. 
However, while applying this to datasets crawled from the web, offensive content should be removed.
% In general, we envisage researchers and practitioners to benefit from our approach especially, when they are not in a capacity to curate expensive annotations. 
In general, we envisage researchers and practitioners to benefit from our approach especially, when expensive human annotations are not available.
More broadly speaking, we also strongly believe that our approach laid blocks for future work on cross-lingually leveraging English skeletons and automatic translations to generate captions for various languages. Hence, when combined with unpaired captioning, this can especially benefit captioning in low resource languages.

%\section{Example Appendix}
%\label{sec:appendix}

%This is an appendix.

\end{document}